\documentclass[draftcls,onecolumn,12pt]{IEEEtran}
\IEEEoverridecommandlockouts
\usepackage{cite}
\usepackage[english]{babel}
\usepackage{float}
\usepackage{stmaryrd} 
\usepackage{nomencl} 
\usepackage{nicefrac}%
\usepackage{stackengine}
\usepackage{color}	
\usepackage{xcolor}
\usepackage{graphicx}			
\usepackage{microtype} 			
\usepackage{cmap}				
\usepackage{epstopdf}           
\usepackage{tikz}
\usepackage{cancel}
\usepackage{enumerate}
\usepackage{amsmath,amsfonts,amssymb,amsthm}
\usepackage{mathrsfs}
\usepackage{mathtools}
\usepackage{hyperref}
\hypersetup{
    colorlinks=true,
    linkcolor=blue,
    filecolor=magenta,      
    urlcolor=blue,
    citecolor=black,
}
\usepackage[capitalize]{cleveref} 

\definecolor{purple}{RGB}{69,22,170}

\newcommand{\trp}{\intercal}
\newcommand{\R}{\mathbb{R}}
\newcommand\iidsim{\stackrel{\mathclap{iid}}{\sim}}

\DeclareMathOperator{\diag}{diag}
\DeclareMathOperator{\blkdiag}{blkdiag}

\DeclareMathOperator{\Ad}{Ad}
\DeclareMathOperator{\ad}{ad}

\usepackage{multirow}
\usepackage{arydshln}
\usepackage{float}
\newtheorem*{remark}{Remark}
\newtheorem{theorem}{Theorem}[section]

\newtheorem{lemma}[theorem]{Lemma}

\title{GNSS/MEMS-INS Integration for Drone Navigation using EKF on Lie Groups}
\author{Marcos R. Fernandes, Giorgio M. Magalhães, Yusef Cáceres, João B. R. do Val%

\thanks{This work was supported in part by the Coordenação de
Aperfeiçoamento de Pessoal de Nível Superior (CAPES), Brazil Finance
Code 001 under Grant 88887.342183/2019-00; in part by the Conselho
Nacional de Desenvolvimento Científico e Tecnológico (CNPq) under
Grant 303352/2018; and in part by FAPESP under Grant 2016/08645-9. The authors are with the School of Electrical and Computer Engineering, University of Campinas -- UNICAMP, Campinas, SP, Brazil. \{marofe,jbosco\}@unicamp.br, }}

\usepackage{caption}
\usepackage{subcaption}
\usepackage{relsize}
\newcommand{\SE}{\textit{SE\/}}

\begin{document}
\maketitle
\begin{abstract}
Building upon the theory of Kalman Filtering on Lie Groups, this paper describes an Extended Kalman Filter and Smoother for Loosely Coupled Integration of GNSS/INS tailored for post-processing applications. The approach employs a dynamic model on a matrix Lie Group that aggregates position, velocity, attitude, and the IMU biases as a single element of a Lie group. The development was motivated by a drone-borne Differential Interferometric SAR (DinSAR) application, which requires high-precision navigation information for short-flight missions using low-cost MEMS sensors. The filter and the Rauch-Tung-Striebel (RTS) smoother are both implemented and validated. The paper also presents a novel algorithm to initialize the heading value as an alternative to gyro-compassing or magnetometer-based alignments. The Mahalanobis Distance and the $\chi^2$-test are employed during the filter update step to address the practical issue of outlier rejection for the GNSS measurements. The paper uses synthetic data to compare classic navigation schemes based on multiplicative quaternions and Euler angles. Finally, real data experiments demonstrate that the Kalman Filter based on Lie Groups performs better DinSAR processing than state-of-the-art commercial software.
\end{abstract}


\section{Introduction}

\IEEEPARstart{R}{emote} sensing applications based on Unmanned Aerial Vehicles (UAV), such
as 3D mapping \cite{Nagai2009} and  Differential Interferometric SAR (DinSAR) \cite{Moreira2019}, require high-fidelity position and attitude information \cite{Gross2012}.

A Strap-down Inertial Navigation System (INS) provides position, velocity, and attitude (PVA) information at high rates but with unqualified errors since it relies on integrating inertial measurements from noisy sensors, such as accelerometers and gyroscopes. On the other hand, a Global Navigation Satellite-based System (GNSS) can provide position and velocity with strict error bounds but at a lower frequency. A Real-Time Kinematic Differential GNSS (RTK-GNSS) could be employed for high-precision applications to achieve centimeter-level precision using code and phase measurements \cite{Langley1998}. A GNSS/INS integrated navigation system combines each sensor's strengths to get better PVA estimates. 

In the literature, different types of GNSS/INS integration have appeared, such as Loosely Coupled (LC), Tightly Coupled (TC), and Ultra‐Tightly Coupled (UTC) \cite{Groves2013}. Due to its simplicity and low computational burden, the LC structure has been vastly used. This integration scheme takes the position provided by the GNSS and combines it with the IMU measurements.

The sensors are inevitably affected by biases, and white noise \cite{Farrell2008}. Besides, with advances in Micro Electromechanical Systems (MEMS) technology, commercial off-the-shelf MEMS-based inertial sensors are available, which are lighter and low-cost but also require adequate noise modeling. In addition, a good alignment process is indispensable to achieving centimeter-level precision.

Over the past decades, algorithms based on the Kalman Filter have been intensively investigated for GNSS/INS integration. A few examples are \cite{Crassidis2006,Edwan2012,Zhang2015}. The classical Extended Kalman Filter (EKF) is the most common GNSS/INS integration technique, and to achieve high-order approximations, unscented or particle filters offer alternatives. They might provide better estimation but require more computational resources than the EKF. For this reason, the EKF remains the reference filter within the aerospace industry, cf. \cite{Barrau2018}.

Several parameterizations apply to represent the attitude of a vehicle, such as Euler Angles, Quaternions, Rotation vectors, and Rotation Matrices, cf. \cite{Groves2013}. The challenge in choosing an adequate representation of attitude is that some representations have singularities or add constraints.

Early applications of Kalman filtering use the three-parameter Euler-angle representation. However, the kinematic equations for Euler angles involve trigonometric functions, which make the model highly non-linear, cf. \cite{Lefferts1982}. Besides, the angles can become undefined for some attitudes, the so-called \emph{gimbal-lock} situation.

Alternatively, quaternions are especially appealing for attitude representation since no singularities are present and the kinematic equation is linear, cf. \cite{Crassidis2007a}. However, the quaternion must satisfy a normalization constraint to represent rotations, which is disregarded by the measurement update step of the standard EKF, cf.  \cite{JohnL.Crassidis2012}. To circumvent such constraint, the Multiplicative Quaternion Extended Kalman Filter (MEKF); see \cite{Lefferts1982,F.Landis2003} was proposed using a multiplicative form of quaternion update that preserves the unity norm.
The advantages of quaternion parameterization have led to its frequent use in attitude determination systems. In fact, since the early 1980s, the quaternion has been the most widely used attitude parameterization, cf. \cite{Crassidis2007a}. Nevertheless, most of these techniques consider models on the Euclidean space and the noises as Gaussian distributed.

Recently, the framework based on Lie Group theory has attracted the attention of sensor fusion communities for rigid-body-related data fusion applications. For instance, a Discrete Extended Kalman Filter on Lie Groups (D-LIE-EKF) appears in \cite{Bourmaud2013a,Bourmaud2016}, which generalizes the usual Kalman Filter framework when the system dynamic or measurement model can be cast as a Lie Group element.

Lie groups can properly model rigid bodies that undergo simultaneous translations and rotations, such as vehicles following curved trajectories that do not suffer from singularities. 
Also, they allow one to implement non-linear estimation tools using linear representations in Lie Algebra. Physical systems modeling, such as the satellite attitude dynamics, fixed-wing unmanned aircraft systems, and multi-rotor flying vehicles, to name a few, are examples that fit as Lie groups naturally.

Classically, the system evolves in the Euclidean space, and the disturbance is taken for granted as additive Gaussian noise. Within D-LIE-EKF, the noise is Gaussian distributed, but acting in the Lie Algebra, which induces a \textit{Concentrated Gaussian distributed} in the Lie Group, cf. \cite{Barfoot2014}.

The robotics community has grown in the awareness that the use of probability distributions defined adequately on Lie groups is an essential tool for pose estimation of rigid bodies. For instance, \cite{Long2012} s that a banana-shape distribution of the unknown position of a differential-drive robot displays a banana-shaped distribution which one can obtain with the aid of the exponential map of the Lie group $\SE(2)$.

In this work, we exploit the generalization of EKF on Lie groups to implement a Loosely Coupled Integration of RTK-GNSS/MEMS-INS for drone-borne remote sensing applications such as DinSAR. The Double Direct Isometries Lie Group $\SE_2(3)$ ( see \cite{Barrau2015})  is adopted to embed the attitude, velocity, and position states combined with the translation group $T(6)$ to accommodate the accelerometer and gyroscope biases. In addition, the RTS smoothing on Lie Groups (see \cite{Bourmaud2015}) is also implemented. The RTS allows one to leverage all information available to obtain the state estimates for each time step, a proposal suitable for post-processing applications requiring higher precision and accuracy.

In the context of navigation, Lie groups have been applied to industrial UAV systems \cite{Barrau2018}, real-time UAV helicopter navigation  \cite{Cui2021}, and also land vehicle navigation \cite{Barczyk2013}.
However, none of these previous studies aimed at developing a complete
navigation system to meet the requirements of aerial mobile mapping, including filtering and smoothing. Most current works focus on land vehicle navigation and filtering solutions only.

Moreover, using simulated data, we show that the proposed filter and smoother outperform conventional quaternion and Euler-based approaches. Secondly, we use a real dataset to show that the Lie Group-based filtering and smoothing yield better performances for DinSAR processing compared to the state-of-the-art commercial software Inertial Explorer\textregistered.

In summary, the main contributions of this paper are three-fold.
\begin{enumerate}[i)]
\item The development of D-LIE-EKF  and RTS smoother for loosely coupled integration of GNSS/INS tailored for post-processing applications that require high precision and accuracy.

\item The proposal of a novel strategy to initialize the heading that does not rely on gyro-compassing or magnetometer, allowing the use of low-cost MEMS inertial sensors.

\item The performance evaluation of  Lie Group-based GNSS/INS integration against conventional methods using simulated data and comparison of DinSAR processing using real dataset against the commercial software Inertial Explorer\textregistered.
\end{enumerate}
To our best knowledge, this is the first implementation of loosely coupled RTK-GNSS/INS integration tailored for post-processing applications, combining Kalman filtering and RTS smoothing on Lie Groups, applied to drone-borne remote sensing applications.

This paper is organized as follows. \cref{sec:lie.theory} briefly introduces
the mathematical concepts for understanding the Kalman Filter algorithm on Lie Groups. \cref{sec:lie.filter} describes the modeling of dynamic systems and random variables on Lie Groups, followed by the description of filtering and smoothing methods. \cref{sec:nav.equations} develops the navigation and sensors model, and \cref{sec:integration} describes the integration scheme. After, experimental evaluations using both simulated and real data appear in \cref{sec:experiments}. Concluding remarks are provided in \cref{sec:conclusion}.

The following notation is adopted: $x_{\beta\alpha}^\gamma$ 
for a vector $x$ represents the coordinates of some kinematic property (position, velocity, etc.) of frame $\alpha$ w.r.t. frame $\beta$, expressed in the frame $\gamma$. The navigation frame is the North-East-Down (NED) local-level frame, abbreviated by the index $n$; the body frame is indicated by $b$, the inertial frame by $i$, whereas the Earth Centered Earth Fixed (ECEF) frame is indexed by $e$.
\section{Filtering and Smoothing on Lie Groups}
\label{sec:lie.filter}

\subsection{Brief Review of Lie Group Theory}
\label{sec:lie.theory}
A Lie group is a mathematical structure that combines the concept of a differentiable manifold with the concept of a group. 
For rigid-body applications, usually, the analysis is restricted to a subgroup of the General Linear Group $GL(n, \mathbb{R})$, also called matrix Lie groups \cite{Hall2015}. The $GL(n, \mathbb{R})$ is the group of all invertible $n \times n$ matrices of real numbers with group operation given by the matrix product. In this paper, we also restrict our consideration to connected and unimodular Lie groups, as it is required for the way the uncertainty on Lie groups is modeled. 

An important structure in the Lie group theory is the \emph{Lie algebra}, a vector space equipped with a bracket product $\llbracket\cdot,\cdot\rrbracket$ called the \emph{Lie bracket}. It is always possible to find a Lie algebra associated with a Lie group\cite{Hall2015}.  For matrix Lie groups, in particular, the associated Lie algebra is the vector space of matrices with Lie bracket given by the commutator $\llbracket X,Y\rrbracket = XY - YX$. 
The importance of the Lie algebra lies in that most of the properties of the Lie group come from properties in the Lie algebra.

The \emph{exponential map} $\exp_G: \mathfrak{g}\to G$, which for a matrix Lie group reduces to the usual matrix exponential function, provides a connection between elements of the Lie group and elements of the Lie algebra.

In general, the exponential map is not bijective; however, it is possible to show that there exist open neighborhoods of the identity element on the Lie group and of the zero element on the Lie algebra, for which the exponential map is a diffeomorphism\footnote{A diffeomorphism is an isomorphism between differentiable manifolds. A differentiable invertible function between manifolds with the differentiable inverse.}. In these open sets, one defines the \emph{logarithm map} $\log_G:G \to \mathfrak{g}$  as the inverse of the exponential map.

Since the Lie algebra is a vector space, it is possible to represent any element as a linear combination of its basis. Therefore, instead of manipulating the elements $X$ as matrices for computing purposes, one can work with the coefficients associated with its basis. With that in mind, the following isomorphisms are defined,
\begin{align}
    [\cdot]^{\vee}_G : \mathfrak{g} &\rightarrow \mathbb{R}^p & [\cdot]_G^{\wedge} : \mathbb{R}^p &\rightarrow \mathfrak{g} \\
    X &\mapsto [X]_G^{\vee} & x &\mapsto [x]_G^{\wedge} \nonumber
\end{align}

For brevity, the following notations are used hereafter,
\begin{align}
\exp_G^\wedge(x):=\exp_G([x]^\wedge_G),\quad
\log_G^\vee(g):=[\log_G(g)]_G^\vee
\end{align}
where $x\in\R^p$, $g\in G$ and when we write $g=\exp_G^\wedge(x)$ we assume that $\log_G^\vee(g)=x$, i.e. we work only on the subsets where $\exp_G(\cdot)$ and $\log_G(\cdot)$ are bijective.

\begin{figure}[H]
    \centering
    \includegraphics[width=0.3\textwidth]{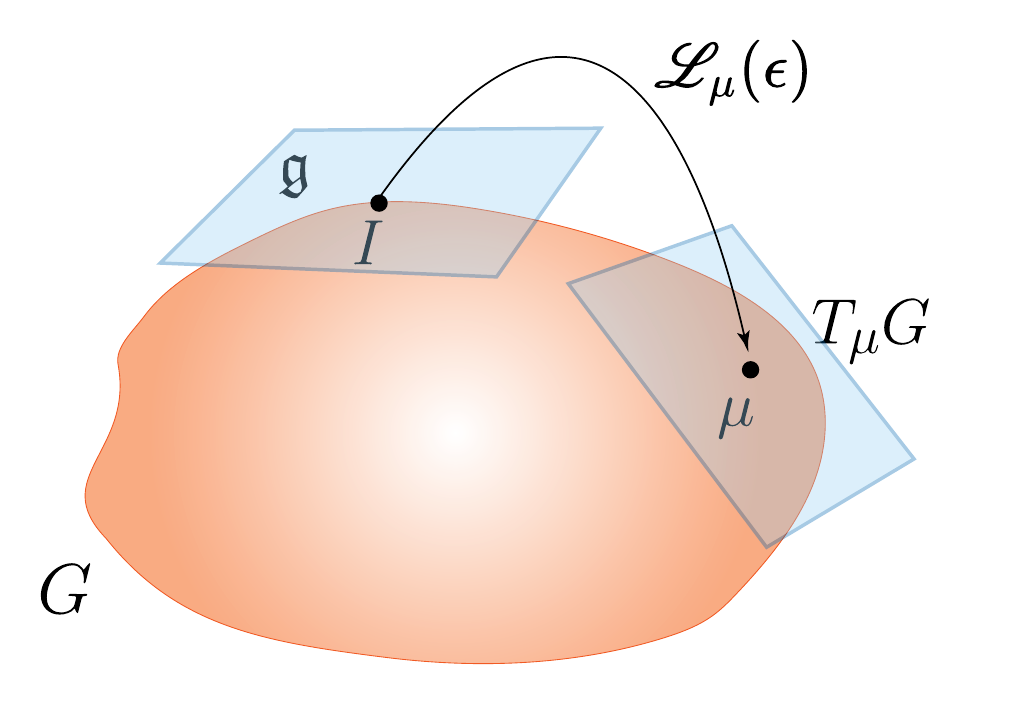}
    \caption{Left Translation.}
    \label{fig:left_action}
\end{figure}

Note that the exponential map can be interpreted as a parameterization for the Lie Group in local coordinates around the identity element. This parameterization can be extended to the neighborhood of any element $\mu\in G$ in a connected Lie group using \emph{Left Translation} as follows,
$
    \mathscr{L}_{\mu}(\epsilon) := \mu \exp_G^{\wedge}(\epsilon),\forall \epsilon\in\mathbb{R}^p
$.
An illustration of the Left Action is depicted in \cref{fig:left_action}.

Since the exponential map is locally a diffeomorphism, there exist open neighborhoods of $\mu\in G$ and $0 \in \mathbb{R}^p$ for which this parameterization is one-to-one.

A Lie group, in general, is not a commutative structure, which can complicate the algebraic manipulations. However, a concept overcomes this issue: \emph{the adjoint representation}. There are two adjoint representations. The first one represents the Lie group on its Lie algebra, i.e., the linear map that takes an element of the Lie group to a linear transformation in the Lie algebra. The Adjoint representation  can be defined as \cite{Chirikjian2012}
\begin{align}
    \Ad_G(g) y=\left[g [y]_G^{\wedge} g^{-1}\right]_G^\vee
\end{align}
where $g \in G$, $y \in \mathbb{R}^p$. Note that $\Ad_G(g)\in\mathbb{R}^{p\times p}$ is a linear transformation that can be applied to any vector $y\in\mathbb{R}^p$.

The second adjoint representation is the representation of the Lie algebra on itself so that each element of the Lie algebra defines a linear transformation in the Lie algebra. This adjoint representation is defined by the Lie bracket \cite{Chirikjian2012} in the form
$
    \left[\ad_G(x) y\right]^{\wedge}_G := \left\llbracket [x]_G^\wedge, [y]_G^\wedge \right\rrbracket
$
where $x,y \in \mathbb{R}^p$. Since the Lie bracket for matrix Lie groups is the commutator operator, we write
\begin{align}
    \ad_G(x)y = \Bigl[[x]_G^\wedge[y]_G^\wedge - [y]_G^\wedge[x]_G^\wedge \Bigr]^{\vee}_G.
\end{align}
Furthermore, in many applications of the Lie group, particularly in filtering and smoothing, one is interested in analyzing the behavior of a Lie group element $g \in G$ as a function of time, and, naturally, its time derivative, $\dot{g}(t)$. From the theory of differential manifolds, $\dot{g}(t)$ is a vector in the vector space tangent to $G$ at the element $g(t)$, i.e. $\dot{g}(t) \in T_{g(t)}G$.

As stated before, working in Lie algebra is a common approach to dealing with Lie groups. It turns out that the tangent space at an arbitrary element of the Lie group is isomorphic to the tangent space at identity by performing a left translation so that $g^{-1} \dot{g} \in \mathfrak{g}$. 
The \textit{right-Jacobian} matrix, following \cite{Chirikjian2012}, is given as
\begin{equation}
    J_r(x) = \sum_{k=0}^{\infty} \frac{(-1)^k}{(k+1)!} \ad_G(x)^k.
\end{equation}
If we consider a local parameterization in the form $g=\mu\exp_G^\wedge(x)$, then the relation of the time variation of $g$ to the time variation of the local coordinates $x$, represented in the Lie algebra, is given by $ g^{-1} \dot{g} = \left[J_r(x) \dot{x}\right]^{\wedge}_G$. Notice that the matrix $J_r(x)$ is the main responsible for relating local coordinates variations to Lie group element variations. 
\subsection{Random Variables on Lie Group}
First, recall that for a random  variable (r.v.) $x$ on the Euclidean space with mean $\mu\in\mathbb{R}^n$ and covariance $P=P^\trp\succ 0$ associate with a pdf $p(x)$, one has%
\begin{subequations}
\begin{align}
    0_{n\times 1}&=\int_{\mathbb{R}^n}(x-\mu)p(x)dx,\\
    P&=\int_{\mathbb{R}^n} (x-\mu)(x-\mu)^\trp p(x)dx
\end{align}
\end{subequations}
where the integration is w.r.t. the Lebesgue measure. 

This definition can be naturally extended to Lie Groups as follows. Given a Matrix Lie Group $G$, a random matrix $X\in G$ with pdf $p(X)$ has mean $\mu\in G$ and covariance $P=P^\trp\succ 0$ defined by
\begin{align}
    0_{n\times n}&=\int_G \log_G^\vee(\mu^{-1}X)p(X)dX,\\
    P&=\int_G \log_G^\vee(\mu^{-1}X)\log_G^\vee(\mu^{-1}X)^\trp p(X)dX.
\end{align}
where the integration is w.r.t. the Haar measure \cite{Chirikjian2012}. 

From this standpoint, the concept of \emph{Concentrated Gaussian Distribution} (CGD) \cite{Bourmaud2015}
is used to define a probability density tailored to matrix Lie groups. The mean is defined in the group, and the covariance is in the Lie algebra. Accordingly, a Gaussian random variable on a matrix Lie group is expressed as follows
\begin{equation}
\label{eq:random.lie}
	X=\mu\exp_G^\wedge(\epsilon),
\end{equation}
and the pdf of $X$ takes the form
\begin{equation}
	p(X):= \alpha \exp\left(-\frac{1}{2}\|\log_G^\vee(\mu^{-1}X)\|_{P^{-1}}^2\right)
\end{equation}
where $\alpha \in \mathbb{R}$ is a normalizing factor to ensure $\int p(X)dX=1$.

\begin{figure}[H]
    \centering
    \includegraphics[width=0.3\textwidth]{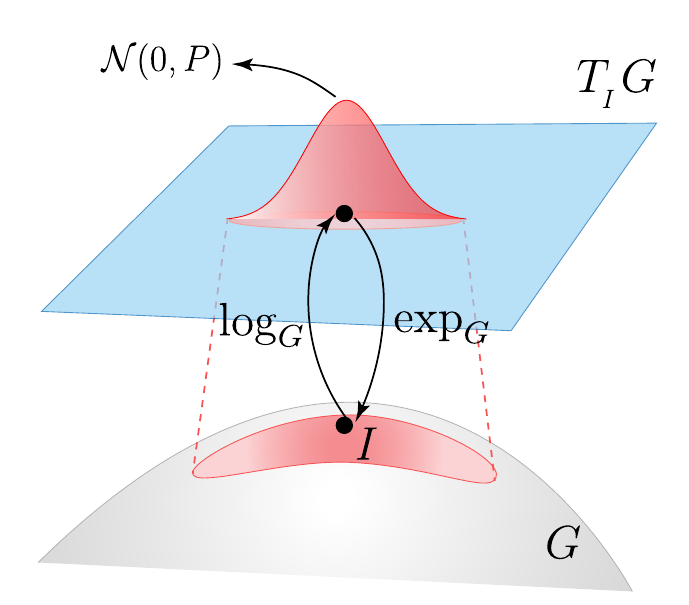}
    \caption{Concentrated Gaussian Distribution in the neighborhood of the identity element. Notice the curved shape of the distribution on the Lie Group.}
    \label{fig:concentrated_gaussian}
\end{figure}

From \cref{eq:random.lie}, $\epsilon=\log_G^\vee(\mu^{-1}X)$ and assuming that $P$ has small eigenvalues then $p(\exp_G^\wedge(\epsilon))$ concentrates around the group identity. With those working assumptions, the distribution of $\epsilon$ in the Lie Algebra becomes the classical Gaussian distribution, i.e. $\epsilon \sim \mathcal{N}(0,P)$. The distribution of $X$ is called a \textit{Concentrated Gaussian Distribution} (CGD) on $G$ and it is denoted by $X \sim \mathscr{N}_G(\mu,P)$. An illustration of the relationship between the neighborhood of the identity element with the Lie Algebra is depicted in \cref{fig:concentrated_gaussian} together with the tangent space and its respective Gaussian distribution.

\subsection{Dynamic System}
Once an r.v. is defined on a Lie Group, a stochastic dynamic system can be modeled such that its states are embedded on a matrix Lie Group $G$. Let $X\in G$ be the system state. Let the stochastic differential equation be expressed by
\begin{equation}
\label{eq:dynamic.system}
dX=X[\Omega(X,u)dt+dW]_G^\wedge
\end{equation}
where $\Omega:G\times\mathbb{R}^m\to \mathbb{R}^p$ is \emph{the left velocity function} and $W$ is a multidimensional Wiener process with covariance $Q_c$, i.e. $W(t_2)-W(t_1)\sim\mathcal{N}(0,(t_2-t_1)Q_c)$ with $t_1<t_2$. Also, consider that measurements are available in discrete time instants in the form 
\begin{align}
\label{eq:measurement.lie}
y_{k+1} &= h(X_k)+\nu_k
\end{align}
where  $y_k\in \mathbb{R}^m$ is the measurement vector, $h:G\to \mathbb{R}^m$ is the measurement function and $\nu_k\iidsim \mathcal{N}(0,R_k)$ is the measurement noise. For small sample time $\Delta t$, the discrete form of the dynamic system \cref{eq:dynamic.system} can be approximated as,
\begin{equation}
    \label{eq:sys.lie}
{X}_{k+1}= X_k\exp_G^\wedge\left( \Omega(X_k,u_k)\Delta t+w_k\right),\\
\end{equation}
where $w_k\iidsim \mathcal{N}(0,Q_k)$ and $Q_k=Q_c\Delta t$.
For convenience, denote $\Omega_k:= \Omega(X_k,u_k)\Delta t$.

\subsection{Kalman Filter on Lie Groups}
Based on the definition of r.v. and stochastic dynamic system on Lie Groups, one can employ the Kalman Filtering framework to generate estimates of a dynamic system state evolving on a Lie Group. The D-EKF on Lie Group \cite{Bourmaud2013a,Bourmaud2016} is presented next.
\begin{lemma}[D-LIE-EKF]
\label{lemma:d.ekf}
The D-EKF on Lie Group is summarized by the following five equations,
\begin{subequations}
\begin{align}
     \hat{X}_{k+1|k}&=\hat{X}_{k|k}\exp_G^\wedge(\hat{\Omega}_k),\\
    P_{k+1|k}&=\mathscr{F}P_{k|k}\mathscr{F}^\trp+J_r(\hat{\Omega}_k)Q_kJ_r(\hat{\Omega}_k)^\trp\\
    K&=P_{k+1|k}\mathscr{H}^\trp(R_{k+1}+\mathscr{H}P_{k+1|k}\mathscr{H}^\trp)^{-1}\\
    \label{eq:filter.update}
    \hat{X}_{k+1|k+1}&=\hat{X}_{k+1|k}\exp_G^\wedge(K(y_{k+1}-h(\hat{X}_{k+1|k})))\\
    P_{k+1|k+1}&=(I-K\mathscr{H})P_{k+1|k}(\bullet)^\trp+ KR_{k+1}K^\trp
\end{align}
where
\begin{align}
    \mathscr{F}&:=\Ad_G(\exp_G^\wedge(-\hat{\Omega}_k))+J_r(\hat{\Omega}_k)\mathscr{C}_k\\
    \mathscr{C}_k&:=\frac{\partial}{\partial \epsilon}[\Omega(\hat{X}_{k|k}\exp_G^\wedge(\epsilon))]\Big|_{\epsilon=0}\\
    \mathscr{H}&:=\frac{\partial}{\partial \epsilon}[h(\hat{X}_{k+1|k}\exp_G^\wedge(\epsilon))]\Big|_{\epsilon=0}.
\end{align}
\end{subequations}
\end{lemma}
\begin{proof}
See \cref{app:d.lie.ekf}.
\end{proof}

\subsection{Smoothing on Lie Groups}
In this section, we provide the required equations to implement the \emph{Rauch-Tung-Striebel} (RTS) smoother on Lie Groups proposed in \cite{Bourmaud2015}.

Unlike the Kalman filter, the smoother uses all available measurements to compute the state estimates using a forward pass, given by the Kalman Filter, followed by a backward pass \cite{Laan2020}. As a consequence, the RTS can only be performed offline.

\begin{lemma}[D-LIE-RTS]
\label{lemma:d.eks}
Given the filter solution $\{\hat{X}_{k|k},P_{k|k}\}_{1:T}$, the Rauch–Tung–Striebel recursion on Lie Groups for $k=T-1,\ldots,1$ are
\begin{subequations}
\begin{align}
 {\hat{X}}_{k+1|k}&= {\hat{X}}_{k|k}\exp_G^\wedge(\hat{\Omega}_k),\\
P_{k+1|k}&=\mathscr{F}P_{k|k}\mathscr{F}^{\trp}+J_r(\hat{\Omega}_k)Q_kJ_r(\hat{\Omega}_k)^\trp,\\
G_k&=P_{k|k}\mathscr{F}^\trp P_{k+1|k}^{-1},\\
\hat{X}_{k}^s&= \hat{X}_{k|k}\exp_G^\wedge(G_k\log_G^\vee( {\hat{X}}_{k+1|k}^{-1} {\hat{X}}_{k+1}^s)),\\
P_{k}^s&=P_{k|k}+G_k(P_{k+1}^s-P_{k+1|k})G_k^\trp.
\end{align}
\end{subequations}
\end{lemma}
\begin{proof}
See \cref{app:rts.lie}.
\end{proof}
\begin{remark}
The D-LIE-EKF presented is in close correspondence with the version presented in \cite{Bourmaud2013a,Bourmaud2015,Bourmaud2016}. However, \cref{eq:posterior.cov} it is in the \emph{Joseph form} for better numeric stability; Moreover, the LIE-RTS is derived using the \emph{left-error} definition instead of the \emph{right-error} definition presented in \cite{Bourmaud2015}.

\end{remark}

\section{Inertial Navigation System}
\label{sec:nav.equations}
\subsection{Navigation Equations}

The kinematic navigation equations (attitude, velocity and position) \cite{Groves2013,Titterton2005}, in continuous time, are given by
\begin{subequations}
\label{eq:nav}
\begin{align}
 \dot{C}_b^e&= C_b^e[\omega_{ib}^b ]^\times-[\omega_{ie}^e ]^\times C_b^e,\\
 \dot{p}^e_{eb}&= v^e_{eb},\\
 \dot{v}^e_{eb}&= C_b^ef_{ib}^b-2 [\omega_{ie}^e]^\times  v^e_{eb}+g^e
    \end{align}
\end{subequations}
where the position $p^e_{eb}=[x_{eb}^e~y_{eb}^e~z_{eb}^e]^\trp$ is coordinated in the \textit{Earth-Centered-Earth-Fixed} (ECEF) frame, and
$$\omega^e_{ie}=[\omega_{e}\cos(L)~0~-\omega_{e}\sin(L)]^\trp$$
where $\omega_e$ is the Earth rotation rate, and $L$ is the latitude; $\omega_{ib}^b,f_{ib}^b\in\mathbb{R}^3$ are the \emph{angular velocity} and \emph{specific force}, respectively. 

The gravity may be obtained through a gravity model.  This paper adopts the model presented in \cite{Rogers2003} due to its simplicity.
\begin{remark}
Since the GNSS navigation solutions are given in the ECEF frame, the INS kinematic model is also defined in the ECEF such that the GNSS measurement can be used to update the trajectory with no coordinate transformation. If a navigation solution coordinate in the local frame is required, one can easily transform from ECEF to NED coordinates (see \cite{Groves2013}).
\end{remark}

\subsection{IMU measurement model}
As pointed out in \cite{Goel2021}, the INS kinematic model in \cref{eq:nav} is exact in the sense that there is no model error or uncertainty. Therefore, the uncertainty in navigation problems comes from the sensors and the local gravity anomalies.

All gyroscopes and accelerometers are subject to errors that limit the accuracy at which angular rotations or specific forces are measured. 
Therefore, modeling sensors for the navigation filter and smoothing becomes essential to achieve reliable results. For this particular case, we consider the following inertial sensor model%
\begin{subequations}
\label{eq:imu}
\begin{align}
 \tilde{\omega}_{ib}^b&= \omega_{ib}^b+ {b}_g+\varepsilon_g, \\
 \tilde{f}_{ib}^b&= f_{ib}^b+ b_a+\varepsilon_a,
\end{align}
\end{subequations}
where ${\varepsilon}_g \iidsim \mathcal{N}(0,\sigma_g^2)$ and ${\varepsilon}_a \iidsim \mathcal{N}(0,\sigma_a^2)$ are the white noise and are related to the \emph{Angular Random Walk} (ARW) and \emph{Velocity Random Walk} parameters of the IMU. In addition, $b_g,b_a$ are the gyroscope and accelerometer biases, respectively. The biases are modeled as \emph{a Random Walk} process in the form
\begin{subequations}
\begin{align}
    db_g&=B_g dW_g,\\
    db_a&=B_adW_a,
\end{align}
\end{subequations}
where $W_g$ and $W_a$ are Wiener processes of appropriate dimensions and $B_g$ and $B_a$ are diffusion matrices associated with the \emph{Bias Instability} of the IMU.

\begin{remark}
It should be noted that in \cref{eq:imu} $\tilde{f}_{ib}^b$ and $\tilde{\omega}_{ib}^b$ are the noisy values of the specific force and angular velocity from the IMU. Moreover, the biases are expressed in the body frame.
\end{remark}

\section{Integration RTK-GNSS/INS}
\label{sec:integration}

\subsection{INS on Lie Group}
In this work, we employ the Double Direct Isometries  $\SE_2(3)$ Lie Group \cite{Barrau2015} for embedding the kinematic states in a compound with a translation group $T(6)$ to accommodate both accelerometer and gyroscope biases. The resulting group structure $G=\SE_2(3)\times T(6)$ is,%
\begin{equation}
\label{eq:group.lie}
    X=\left[\begin{array}{c;{2pt/2pt}r}
    \begin{matrix}
    C_b^e & v^e & p^e \\
    0_{1\times3} & 1 & 0 \\
    0_{1\times3} & 0 & 1 \end{matrix} & 0_{5\times 7}  \\\hdashline[2pt/2pt]
    0_{7\times5} & \begin{matrix} I_{6\times6} & b\\
 0_{1\times6} & 1
      \end{matrix}
    \end{array}\right]_{12\times 12}
\end{equation}
where $b=\left[\begin{smallmatrix}b_a\\b_g\end{smallmatrix}\right] \in \R^6$.
\begin{lemma}
The velocity function $\Omega:G\times \mathbb{R}^6\to\mathbb{R}^{15}$ associated with the navigation equations \cref{eq:nav} embedded into the Lie Group from \cref{eq:group.lie} is given by
\begin{equation}
\label{eq:left.velocity}
    \Omega(X,u)=
    \begin{bmatrix}
    \tilde{\omega}_{ib}^b-b_g-\omega_{ie}^b\\
    \tilde{f}_{ib}^b-b_a-2C_e^b[\omega_{ie}^e]^\times v_{eb}^e+C_e^bg^e\\
    C_e^bv_{eb}^e\\
    0\\
    0
    \end{bmatrix}
\end{equation}
where $u=\left[\begin{smallmatrix}\tilde{f}_{ib}^b\\\tilde{\omega}_{ib}^b\end{smallmatrix}\right]\in \R^6$ is the noisy input measurement from the IMU. In addition, the process noise is given by
\begin{equation}
\label{eq:process.noise}
dW=\begin{bmatrix}
-\sigma_adW_1\\
-\sigma_gdW_2\\
0\\
B_adW_a\\
B_gdW_g
\end{bmatrix}=\Gamma d\tilde{W}
\end{equation}
where $d\tilde{W}=[dW_1^\trp~dW_2^\trp~dW_a^\trp~dW^\trp_g]^\trp$ such that $\tilde{W}$ is a standard $12$-dim Wiener process.
\end{lemma}
\begin{proof}
From \cref{eq:group.lie,eq:nav}, one gets
\begin{equation*}
    \begin{aligned}
   X^{-1}dX&=\left[\begin{array}{c;{2pt/2pt}r}
    \begin{matrix}
   C_e^b dC_b^e & C_e^b dv_{eb}^e & C_e^b dp_{eb}^e \\
    0 & 0 & 0 \\
    0 & 0 & 0\\\end{matrix} & 0_{5\times 7}  \\\hdashline[2pt/2pt]
    0_{7\times5} & \begin{matrix} 0_{6\times6} & db\\
 0_{1\times6} & 0
      \end{matrix}
    \end{array}\right]_{12\times 12}
    \\&=[\Omega(X,u)dt+dW]^\wedge_G.
    \end{aligned}
\end{equation*}
Which implies \cref{eq:left.velocity,eq:process.noise}.
\end{proof}
\subsection{GNSS Measurement Model}
Assuming that the rover GNSS antenna is rigidly fixed relative to the IMU and the lever-arm from the IMU to the GNSS antenna phase center is expressed in the body frame as $l^b$, the GNSS measurements $p^\text{GNSS}=[x~y~z]^\trp$ are modeled using
\begin{equation}
 p^\text{GNSS}=p^e_{eb}+C_b^e{l}^b+ \nu,
\end{equation}
where ${\nu} \iidsim \mathcal{N}(0, R)$ stands for the GNSS noise, assumed to be uncorrelated white noise with covariance $R=\diag(\sigma_x^2,\sigma_y^2,\sigma_z^2)$. 
\subsection{Outlier Rejection}
Outliers are spurious data that differ dramatically from the statistical distribution, leading to
erroneous behavior of the filtering algorithm. When adopting RTK-GNSS as measurement, outliers are inevitably present due to satellite blockage, multi-path noise, or cycle slip.
To make the proposed algorithm robust against GNSS outliers, the Mahalanobis Distance combined with the $\chi^2$-test is employed to define a rejection scheme.

\begin{figure}[H]
    \centering
    \includegraphics[width=0.22\textwidth]{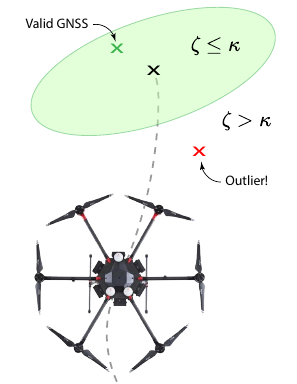}
    \caption{GNSS Outlier Rejection using Mahalanobis Distance.}
    \label{fig:outlier.rejection}
\end{figure}

Define the Normalized Residue Squared (NRS) as
\begin{equation}
\zeta_k:=\|y_k-\hat{y}_k\|_{\Xi_k^{-1}}^2   
\end{equation}
where $\Xi_k=R_k+\mathscr{H}_kP_{k+1|k}\mathscr{H}_k^\trp$. Assuming that the NRS obeys a Chi-square distribution with 3 degrees of freedom, i.e., $\zeta_k\sim \chi^2_3$, a valid GNSS is declared if $\zeta_k\le \kappa$ where $\kappa$ is a threshold level. Figure \ref{fig:outlier.rejection} illustrate the region of valid GNSS (green shadow) where the NRS is less than the specified threshold.
However, similar to the method shown in \cite{Chiella2019}, instead of completely discarding a possible outlier when the threshold is exceeded, another approach is to mitigate the influence of respective GNSS measurement by weighting the filter innovation with the following factor,
\begin{equation}
    \gamma=\min\left(1,\frac{\kappa}{\zeta}\right).
\end{equation}
Accordingly, the update equation \cref{eq:filter.update} becomes
\begin{equation}
    \hat{X}_{k+1|k+1}:=\hat{X}_{k+1|k}\exp_G^\wedge(\gamma K(y_{k+1}-h(\hat{X}_{k+1|k}))).
\end{equation}
Observe that for $\zeta\le \kappa$ one has $\gamma=1$. Thus, the filter innovation is unaltered, but for $\zeta>\kappa$, $\gamma$ becomes less than one, scaling down the filter innovation.

\subsection{Alignment and Initial Conditions}
\label{sec:alignment}
A critical factor for achieving accurate navigation is the initialization of the
inertial navigation system.

Assuming that the INS is stationary before the take-off, the initial velocity can be set to zero, and the initial position can be obtained as the average of GNSS measurements during a specified time window. Similarly, the initial values for the gyroscope biases can be estimated using the average of its measurements during the stationary period.

Moreover, the \emph{leveling method} \cite{Groves2013} can be used to initialize the initial pitch ($\theta_0$) and roll ($\phi_0$) angles as well. The principle of leveling is that when the INS is stationary, the accelerometer triad detects only the gravity acceleration. Hence, the initial pitch and roll angles can be obtained as follows, 
\begin{subequations}
\label{eq:leveling}
\begin{align}
	\theta_0&=\arctan\left(\frac{\bar f_{ib,x}^b}{\sqrt{(\bar f_{ib,y}^b)^2+(\bar f_{ib,z}^b)^2}}\right),\\
	\phi_0&=\arctan_2(-\bar f_{ib,y}^b,-\bar f_{ib,z}^b)
\end{align}
\end{subequations}
where $\bar{f}_{ib,x}^b, \bar{f}_{ib,y}^b, \bar{f}_{ib,z}^b$ are the average accelerometer output during a time-window. Note that the four-quadrant arc tangent function should be used for roll. The accuracy of \cref{eq:leveling} is determined by the accelerometer biases \cite{Titterton2005}. 

The gyro-compassing method or a magnetometer compass may be used for the initial heading. However, an accurate initialization of heading requires expensive aviation-grade gyroscope sensors capable of measuring the Earth's rotation rate. Also, for some applications, magnetometers cannot be used due to the presence of magnetic interferences.
Therefore, it is desirable to have a method that  relies solely on low-cost MEMs to obtain a satisfactory initialization of the heading value. With that in mind, we propose an optimization-based method  as described next.

\subsection{Heading Alignment}
\label{sec:heading.align}
To provide a reliable initialization of the heading angle ($\psi_0$) using only a low-cost MEM sensor, we employ the Bayesian parametric estimation scheme described at \cite[cap.12]{Sarkka2013}. It consists of evaluating the posterior distribution $p(\psi_0|y_{1:N})$ and taking the most likely value for the initial heading $\psi_0$.

For this purpose, note that
\begin{equation}
    p(\psi_0|y_{1:N})\propto p(y_{1:N}|\psi_0)p(\psi_0)
\end{equation}
where $p(\psi_0)$ is some prior distribution for the initial heading. For simplicity, we consider $p(\psi_0)=\mathcal{N}(0,\sigma_\psi^2)$.
Moreover, $p(y_{1:N}|\psi_0)$ can be factored in the form 
\begin{equation}
p(y_{1:N}|\psi_0)=\prod_{k=1}^N p(y_k|y_{1:k-1},\psi_0).    
\end{equation}
We assume that the marginal measurement distribution $p(y_k|y_{1:k-1},\psi_0)$ is a Gaussian distribution in the form 
\begin{equation}
    p(y_k|y_{1:k-1},\psi_0)=\mathcal{N}(h(\hat{X}_{k|k-1}^{\psi_0}),R_k+\mathscr{H}P_{k|k-1}^{\psi_0}\mathscr{H}^\trp)
\end{equation}
where $\hat{X}_{k|k-1}^{\psi_0}$ and $P_{k|k-1}^{\psi_0}$ come from the filtering solution with some fixed $\psi_0$ value.
Accordingly, the most likely value for the initial heading can be found by solving $\min_{\psi_0} -\log(p(\psi_0|y_{1:N}))$. 

Let $\varphi(\psi_0)=-\log(p(\psi_0|y_{1:N}))$ then one has
\begin{multline}
    \varphi(\psi_0)=\sum_{k=1}^N -\log(p(y_k|y_{1:k-1},\psi_0))-\log(p(\psi_0))\\ 
    =\sum_{k=1}^N\frac{1}{2}\log(2\pi|S_k^{\psi_0}|)+\frac{1}{2}\|z_k^{\psi_0}\|^2_{(S_k^{\psi_0})^{-1}}+\frac{1}{2\sigma^2_\psi}\|\psi_0\|^2
    \label{eq:energy}
\end{multline}
with $S_k^{\psi_0}=R_k+\mathscr{H}P_{k|k-1}\mathscr{H}^\trp$ and $z_k^{\psi_0}=y_k-h(\hat{X}_{k|k-1}^{\psi_0})$.
\begin{remark}
Note that, for each value of $\psi_0$, \cref{eq:energy} provides the respective log-likelihood up to a constant as $p(\psi_0|y_{1:N})\propto \exp(-\varphi(\psi_0))$. Therefore, for a sufficient number of evaluations of different values for $\psi_0$, one is able to build the distribution $p(\psi_0|y_{1:N})$ using the filtering solution.
\end{remark}

Furthermore, we propose the following approximation $p(\psi_0|y_{1:N})\approx \mathcal{N}(\psi_0^*,\sigma^2_{\psi^*})$. This implies that the log-likelihood \cref{eq:energy} is approximated quadratic w.r.t $\psi_0$; thus, one can obtain the approximated log-likelihood function by evaluating three distinct points.

In summary, the proposed heading alignment scheme consists of three independent Kalman filter evaluations  such that each one has different heading values $[\psi_1~\psi_2~\psi_3]$. After running the filtering, three samples from the log-likelihood are obtained. Thereafter, a parabola $c=m_1\psi^2+m_2\psi+m_3$ is fitted to the samples solving $A m=c$ for $m$ where
\begin{equation}
    A=\begin{bmatrix}
    \psi_1^2 & \psi_1 & 1\\
    \psi_2^2 & \psi_2 & 1\\
    \psi_3^2 & \psi_3 & 1
    \end{bmatrix},c=\begin{bmatrix}
    c_1\\
    c_2\\
    c_3
    \end{bmatrix},m=\begin{bmatrix}
    m_1\\
    m_2\\
    m_3
    \end{bmatrix}.
\end{equation}
The best estimate of the initial heading is then chosen as the value that minimizes the fitted parabola,
$
    \psi_0^*:=-\frac{m_2}{2m_1}
$.

\begin{figure}
    \centering
    \includegraphics[clip,trim=3cm 10cm 3cm 10cm,width=0.4\textwidth]{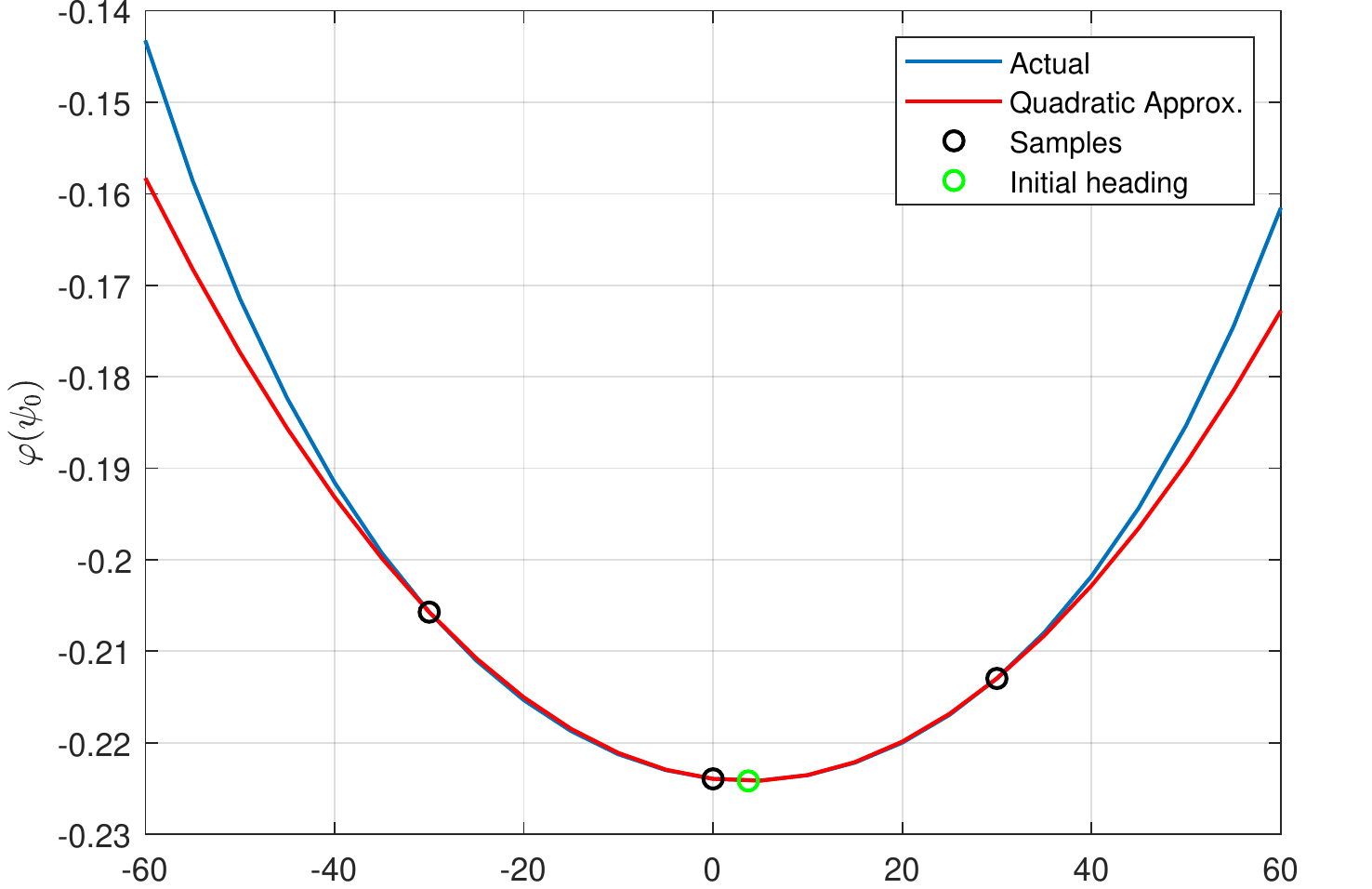}
    \caption{Example of the proposed Optimization-based heading alignment using $-30^\circ,0^\circ,+30^\circ$ guesses for the initial heading. The blue curve indicates the actual log-likelihood function and the red curve is the proposed approximation.}
    \label{fig:parabola}
\end{figure}

Figure \ref{fig:parabola} shows an example of the proposed optimization-based heading alignment with $\psi_1=-30^\circ$, $\psi_2=0^\circ$ and $\psi_3=30^\circ$ applied to a real dataset. For this case, the best initial heading was $\psi_0^*=3.72^\circ$. The actual log-likelihood (blue curve) was computed by a grid of values for $\psi_0$ from $-60^\circ$ to $+60^\circ$ with $5^\circ$ increment. Notice that the blue curve is close to the red curve (quadratic approximation) near its minimum.
In \cref{subsec:heading} a numeric experiment using synthetic data demonstrates the performance of the proposed heading alignment strategy.

\subsection{GNSS/INS integration scheme}
\begin{figure}
    \centering
    \includegraphics[width=0.25\textwidth]{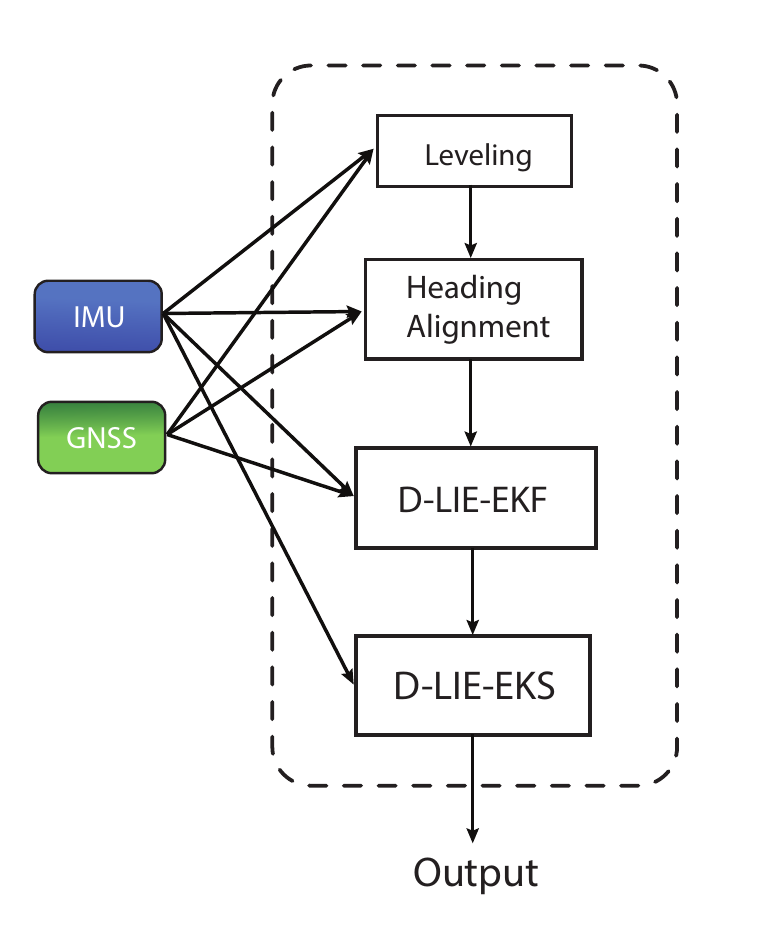}
    \caption{Proposed Loosely Coupled GNSS/INS integration using Lie Group for post-processing applications.}
    \label{fig:lc.diagram}
\end{figure}

The proposed scheme for loosely coupled GNSS/INS integration using Lie Group consists of four steps. First, the accelerometer data from the IMU during an initial stationary period is used to perform leveling and obtain initial values for pitch and roll. After that, the heading alignment method described in \ref{sec:heading.align} is applied to obtain the initial heading value. Next, the D-LIE-EKF algorithm is performed to obtain a filtered solution. Finally, the smoothing method using D-LIE-EKS is used to generate the final output. Figure \ref{fig:lc.diagram} illustrate these four steps.

\section{Data Experiments}
\label{sec:experiments}
\subsection{Settings}
\label{subsec:settings}
This work was motivated by the drone-borne DinSAR application described in \cite{Moreira2019,Luebeck2020}. This application requires high-fidelity PVA information to provide reliable interferometric results.

\begin{figure}
    \centering
    \includegraphics[width=0.35\textwidth]{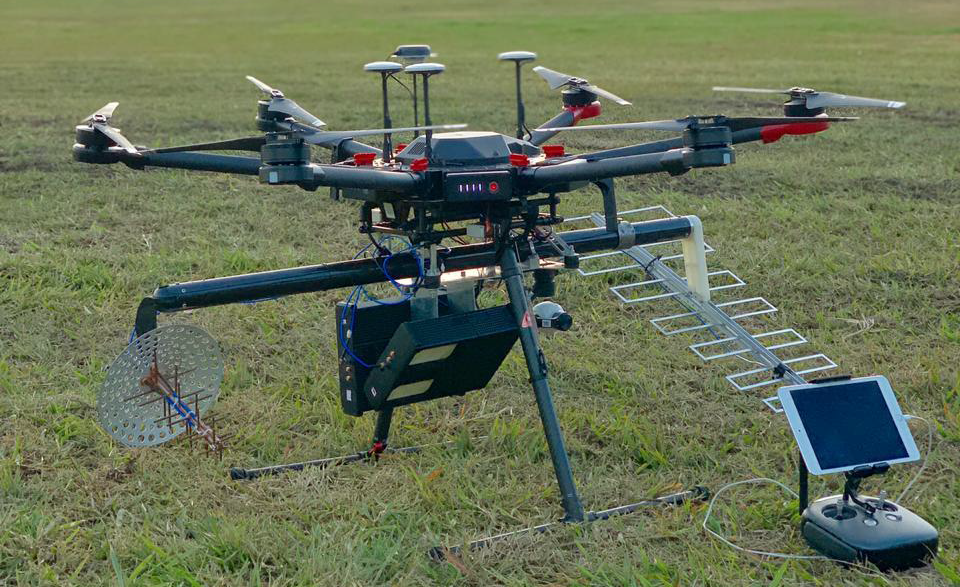}
    \caption{Drone-borne DInSAR system \cite{Moreira2019}.}
    \label{fig:hexacopter}
\end{figure}

The drone with the radar system is shown in Figure \ref{fig:hexacopter}. It consists of a DJI\textregistered~ Matrice 600-Pro equipped with a radar system for remote sensing. 
The native onboard navigation system of the drone is not used. An independent INS is mounted exclusively to provide PVA information for the radar system, consisting of a 6-DOF IMU ADIS16495 from Analog Devices\textregistered. 
The IMU is rigidly mounted on the radar antenna so that the attitude measurement from the IMU can be easily transformed into the orientation information of the radar. 

Moreover, the u-blox\textregistered~ ZED F9P GNSS system is used to provide raw code and phase measurements that are post-processed using the open-source package RTKlib \cite{TomojiTakasu2009}.
\subsection{Simulation}
The following approach was pursued to compare and validate the proposed filter and smoothing scheme. First, a real dataset was processed using the commercial software Inertial Explorer\textregistered~ for GNSS/INS integration with centimeter-level accuracy. The resulting solution was adopted as a reference trajectory. Afterward, an inverse strap-down mechanization similar to \cite{Gonzalez2015a} but adapted to the Lie Group model is implemented to emulate perfect measurements.  Finally, perturbations reflecting characteristic errors of the real sensors were added to the emulated measurements and used to test the filtering and smoothing algorithms.

More specifically, let $\{C_b^e(k),p^e(k),v^e(k)\}_{k=1:N}$ be the reference trajectory generated with a sample rate $F_s$. For each time instant $k$, an $\SE_2(3)$ element is built to represent the system state in the form,
\begin{equation}
    S_k=\begin{bmatrix}
    C_b^e(k) & v^e(k) & p^e(k)\\ 
    0 & I
    \end{bmatrix}.
\end{equation}
Thereafter, the left-velocity vector is computed using
\begin{equation}
  \Omega=\frac{\log_G^\vee(S_k^{-1}S_{k+1})}{\Delta t} 
 =\begin{bmatrix}\Omega_\omega\\\Omega_f\end{bmatrix}.
\end{equation}
From \cref{eq:left.velocity}, one obtains the angular velocity and the specific force as follows,
\begin{align*}
    \omega_{ib}^b(k)&=\Omega_\omega+\omega_{ie}^e,\\
    f_{ib}^b(k)&=\Omega_f+2C_e^b(k)[\omega_{ie}^e]^\times v_{eb}^e(k)-C_e^b(k)g^e.
\end{align*}

Finally, the IMU artificial noises are added to form the simulated IMU measurements
\begin{align*}
    \tilde{\omega}_{ib}^b(k)&=\omega_{ib}^i(k)+b_g(k)+N_gw(k),\\
    \tilde{f}_{ib}^b(k)&=f_{ib}^i(k)+b_a(k)+N_aw(k)
\end{align*}
where $w_a,w_g\sim\mathcal{N}(0,I)$ and $N_a$, $N_g$ are the VRW and ARW parameters of the emulated IMU. Besides, $b_a,b_g$ are the accelerometer and gyroscope biases simulated using the Ornstein-Uhlenbeck process as follows,
\begin{align*}
    b_a(k+1)&=b_a(k)+\tau_a(\beta_a-b_a(k))\Delta t +B_a\sqrt{\Delta t}w_{ba}(k),\\
    b_g(k+1)&=b_g(k)+\tau_g(\beta_g-b_g(k))\Delta t +B_g\sqrt{\Delta t}w_{bg}(k)
\end{align*}
where $\tau_a,\tau_b$ are the biases' correlation time, $\beta_a,\beta_b$ are the turn-on constant biases, $w_{ba},w_{bg}\sim\mathcal{N}(0,I)$ and $B_a,B_g$ are parameters to influence the bias instability. 

\begin{remark}
It is noteworthy that, unlike the Random Walk process, the Ornstein-Uhlenbeck process is a \emph{mean-reverting} process, which seems more appropriate to simulate the IMU biases.
\end{remark}

The GNSS data is generated by down-sampling the reference trajectory to $1Hz$ and adding the lever-arm component together with a 3-dimensional white noise as follows,
\begin{equation}
    y(n)=p^e(nF_s)+C_b^e(nF_s)l^b+\varepsilon(n)
\end{equation}
for $n=0,1,\ldots,\lfloor\nicefrac{N}{F_s}\rfloor$, $\varepsilon\sim\mathcal{N}(0,R)$.

\begin{figure*}
\centering
     \begin{subfigure}[b]{0.3\textwidth}
         \centering
         \includegraphics[clip,trim=4cm 9cm 4cm 8.5cm,width=\textwidth]{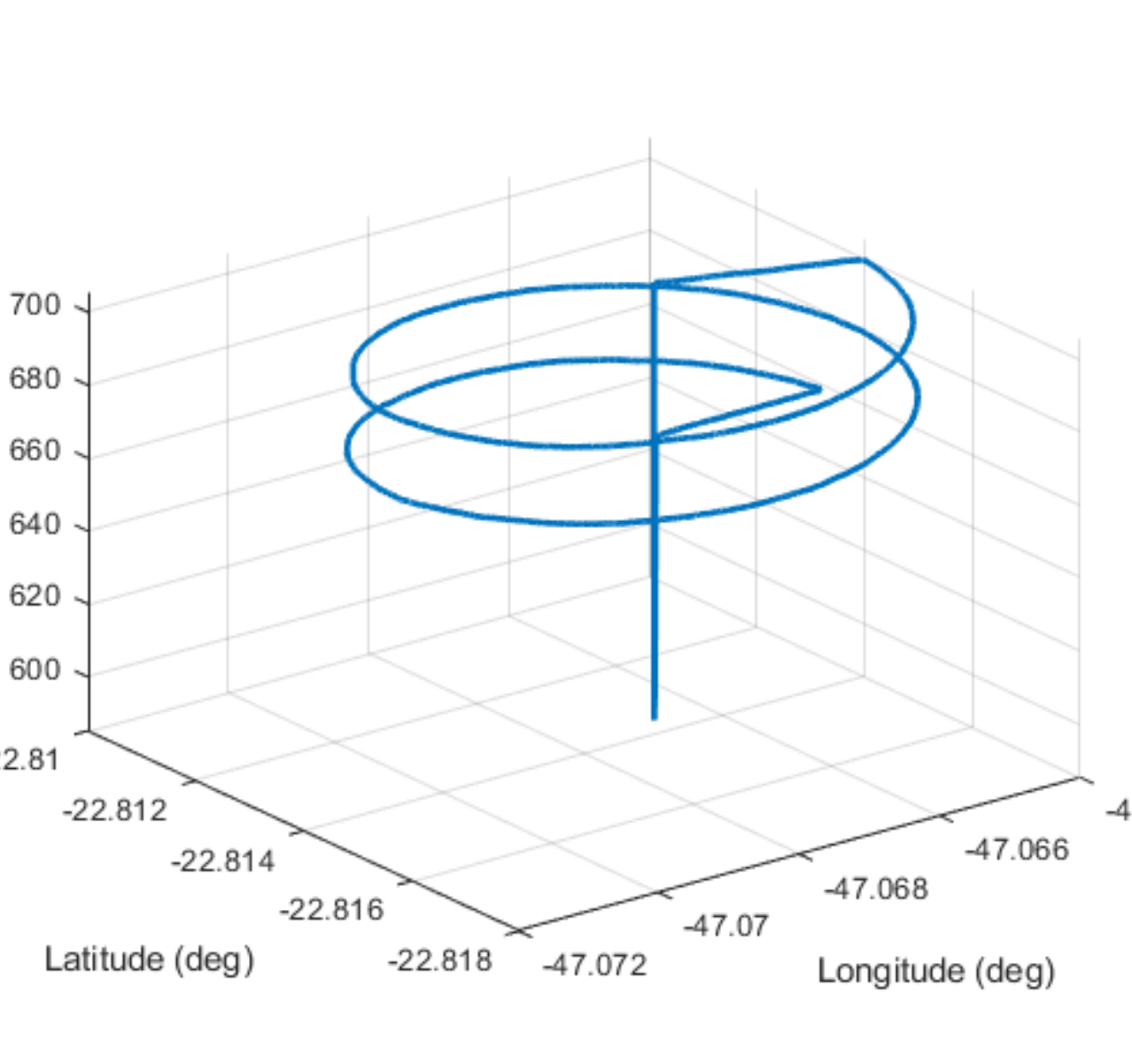}
         \caption{Helicoidal.}
         \label{fig:helicoidal}
     \end{subfigure}
     \hfill
     \begin{subfigure}[b]{0.3\textwidth}
         \centering
         \includegraphics[clip,trim=4cm 9cm 4cm 8.5cm,width=.9\textwidth]{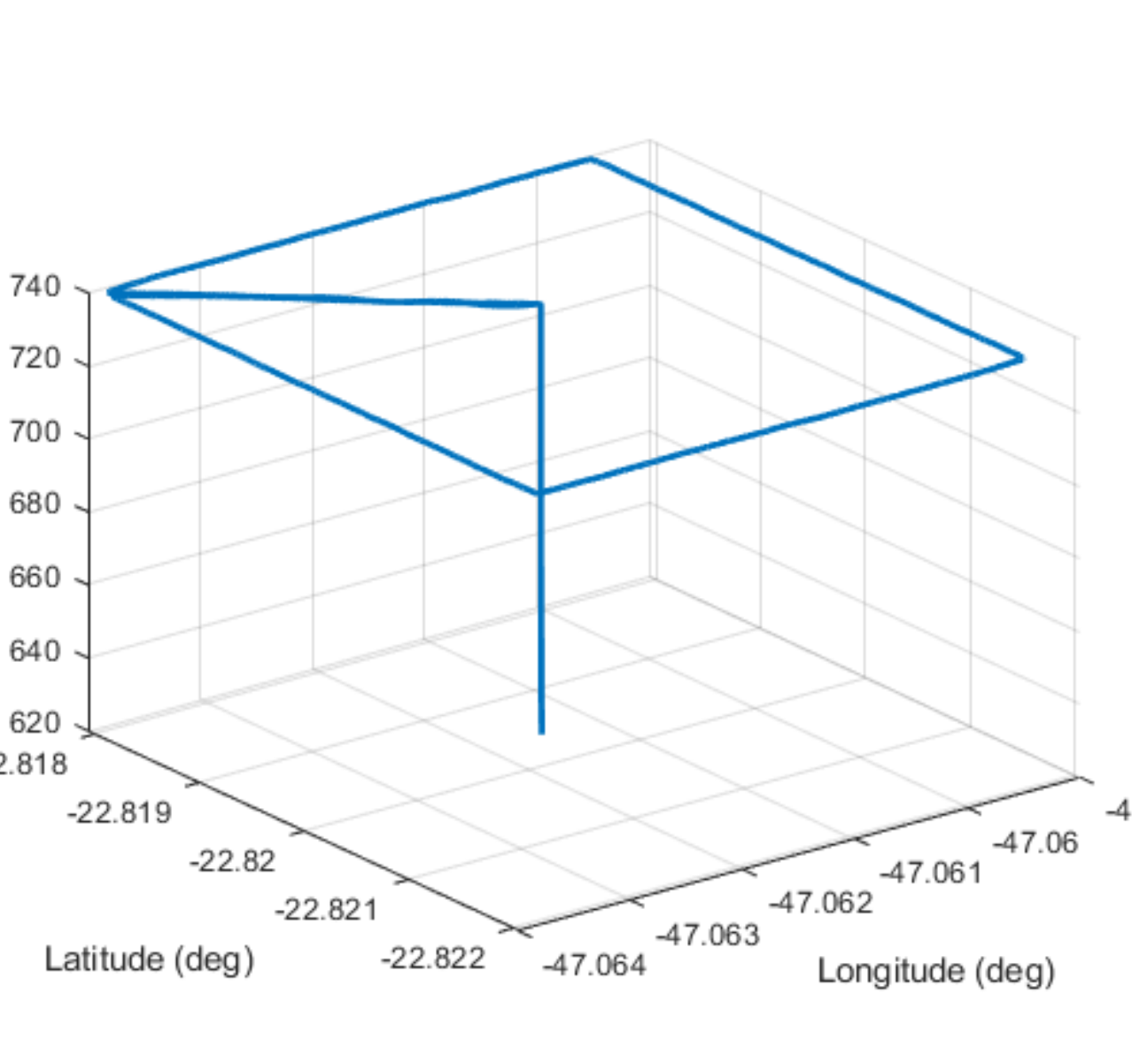}
         \caption{Rectangular.}
         \label{fig:rectangular}
     \end{subfigure}
     \hfill
     \begin{subfigure}[b]{0.3\textwidth}
         \centering
         \includegraphics[clip,trim=4cm 9cm 4cm 8.5cm,width=\textwidth]{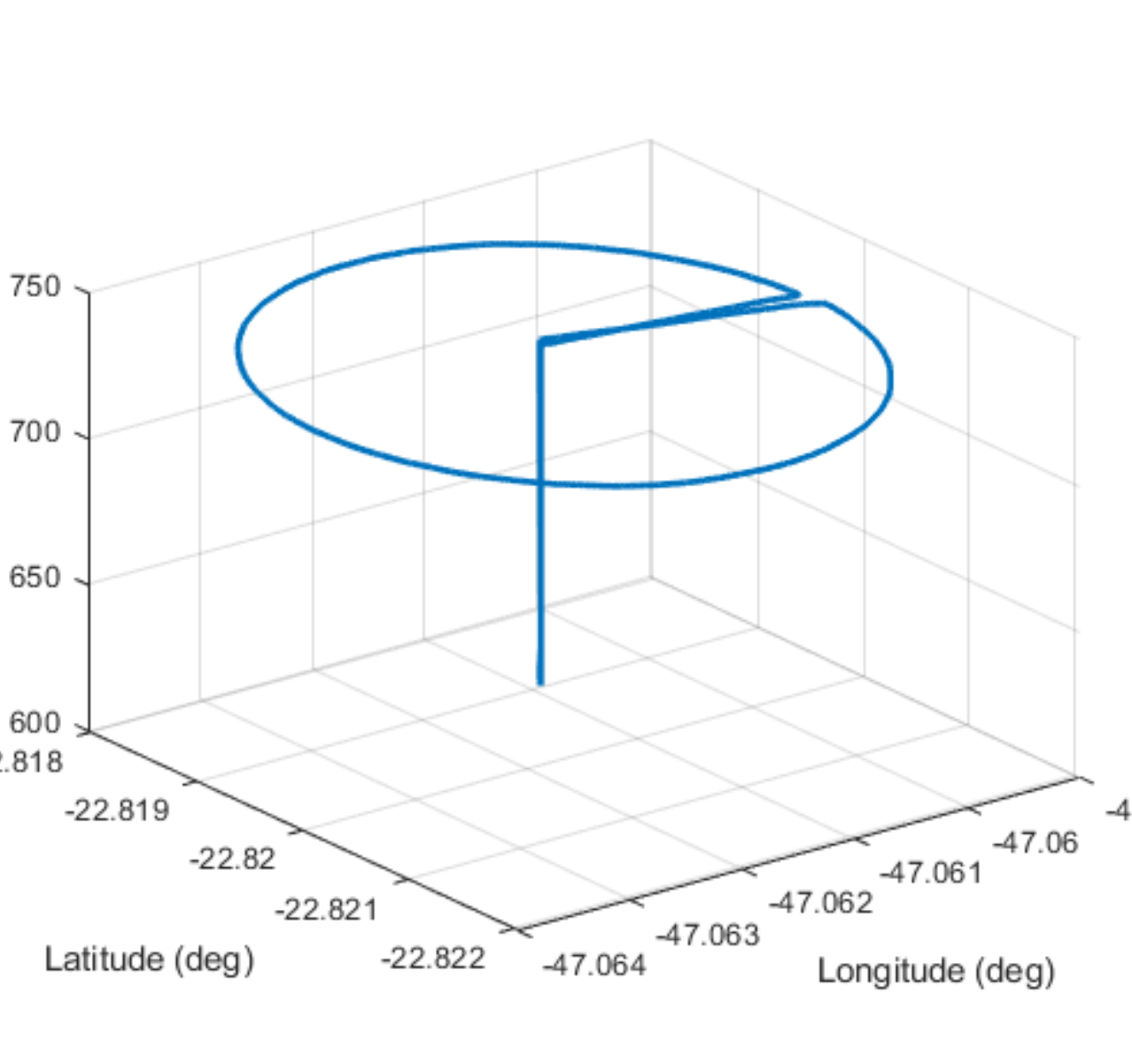}
         \caption{Circular.}
         \label{fig:circular}
     \end{subfigure}
        \caption{Example of three flight scenarios.}
        \label{fig:flight.profiles}
\end{figure*}

The parameters used for all simulations were chosen to match the ADIS16495 data sheet and the centimeter-level precision of RTK-GNSS and are given in \cref{tab:sim.param}.
\begin{table}\caption{Simulation parameters.}
    \centering
    \begin{tabular}{c|c}
    Parameters & Value\\ 
    \hline\\
         $N_a$ & $0.008 \nicefrac{(m/s)}{\sqrt{h}}$  \\
         $N_g$ &  $0.09^\circ/\sqrt{h}$\\
         $B_a$ & $3.2\mu g$\\
         $B_g$ & $0.8^\circ/h$\\
         $\beta_a$ & $500\mu g$\\ 
         $\beta_g$ & $10^\circ/h$\\
         $\tau_a$ & 1s\\ 
         $\tau_g$ & 1s\\
         $\sigma_x$ & $0.01m$\\
         $\sigma_y$ & $0.01m$ \\ $\sigma_z$ & $0.03m$
    \end{tabular}
       \label{tab:sim.param}
\end{table}%
The covariance matrix $P_0$ is formed as a diagonal matrix whose individual elements are chosen according to each state's expected $99.7\%$ confidence interval (three standard deviations), as follows
\begin{multline*}
    P_0=\diag\Bigl(\bigl(\frac{1}{3}^\circ\bigr)^2,\bigl(\frac{1}{3}^\circ\bigr)^2,\bigl(\frac{5}{3}^\circ\bigr)^2,\\ \bigl(\frac{0.001}{3}m/s\bigr)^2,\bigl(\frac{0.001}{3} m/s\bigr)^2,\bigl(\frac{0.001}{3} m/s\bigr)^2,\\
    \bigl(\frac{0.1}{3}m \bigr)^2,\bigl(\frac{0.1}{3}m \bigr)^2,\bigl(\frac{0.1}{3}m \bigr)^2,\\
    \bigl(\frac{1}{3}mg\bigr)^2,\bigl(\frac{1}{3}mg\bigr)^2,\bigl(\frac{1}{3}mg\bigr)^2,\\
    \bigl(\frac{15}{3}\nicefrac{\deg}{s}\bigr)^2,\bigl(\frac{15}{3}\nicefrac{\deg}{s}\bigr)^2,(\frac{15}{3}\nicefrac{\deg}{s}\bigr)^2\Bigr).
\end{multline*}%
\subsection{Heading Alignment Performance}
\label{subsec:heading}
To analyze the accuracy of the proposed heading alignment method described in \cref{sec:heading.align}, we perform a set of experiments with synthetic data using different initial heading values. The optimization-based method is applied for each experiment, and the best estimate of the initial heading is obtained. 
\begin{figure}
    \centering
    \includegraphics[clip,trim=3.5cm 9cm 4cm 9cm,width=0.45\textwidth]{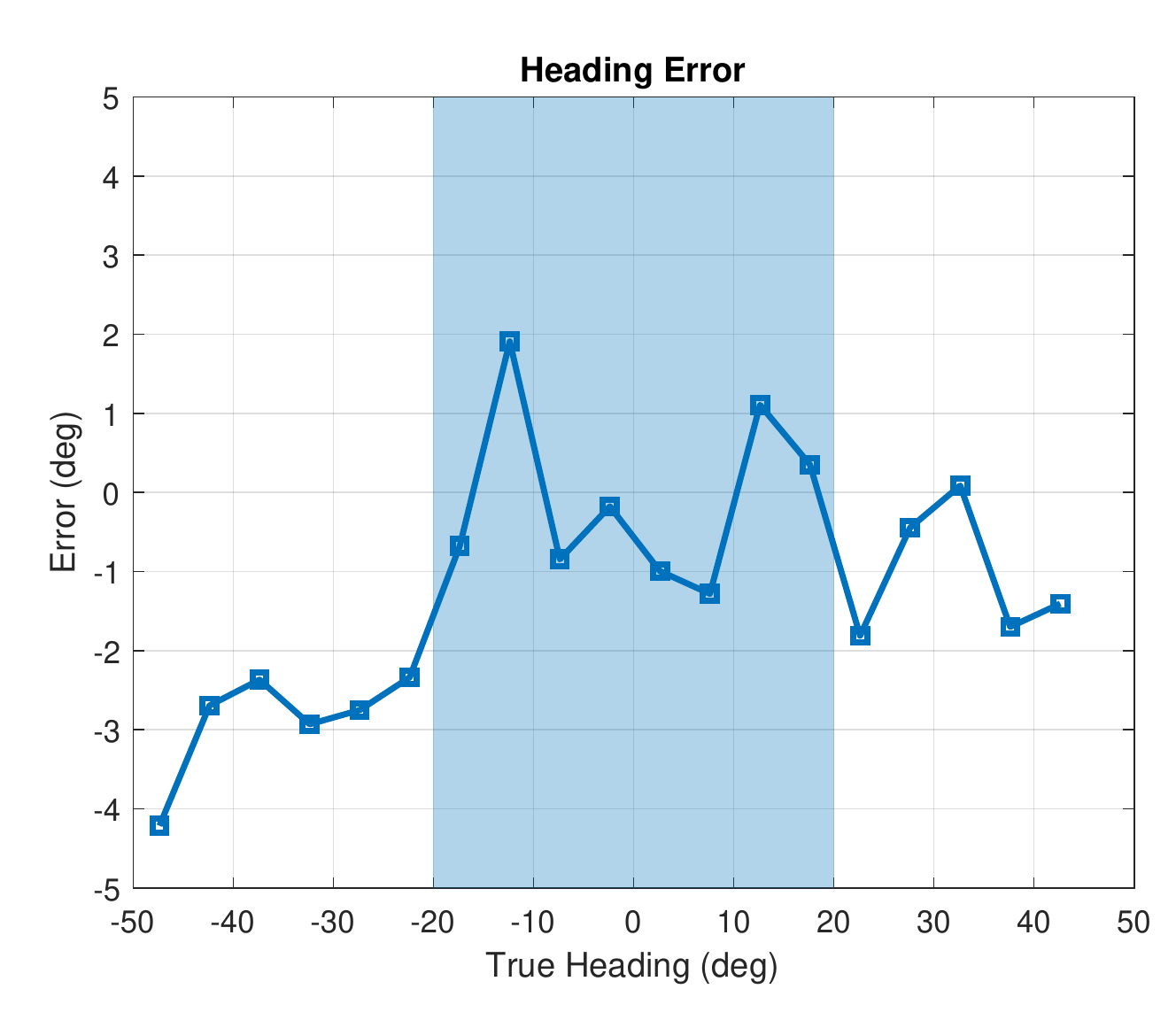}
    \caption{Heading alignment error.}
    \label{fig:heading.error}
\end{figure}%
The experiments support that the proposed heading alignment can achieve an error less than $2^\circ$ when the true heading is between $-20^\circ$ and $+20^\circ$ as shown in Figure \ref{fig:heading.error}.
\subsection{Navigation performance}
Three flight scenarios were simulated to evaluate the proposed processing scheme's performance: helicoidal, rectangular, and circular. These profiles are shown in Figure \ref{fig:flight.profiles}.

For each scenario, the following algorithms were implemented: Lie Group-based filter (D-LIE-EKF) and smoother (D-LIE-EKS); The Multiplicative Quaternion-based filter (MEKF) and smoother (MEKS); Euler-based filter (EULER-EKF) and smoother (EULER-EKS).

Thereafter, 100 Monte Carlo realizations were performed, and each algorithm's respective RMSE was computed.  All algorithms are initialized with the same parameters for a fair comparison. Table \ref{tab:mse.filter} summarizes the performance for each online processing part scenario that only uses filtering. Table \ref{tab:mse.smoother} shows the off-line performance after applying the respective smoother algorithm.

\begin{table}
\centering
\caption{RMSE comparison for 100 Monte Carlo simulations using filter only.}
\resizebox{\columnwidth}{!}{
    
    \begin{tabular}{c||c|c|c||c|c|c||c|c|c}
    & \multicolumn{3}{c||}{Helicoidal} & \multicolumn{3}{c||}{Rectangular}& \multicolumn{3}{c}{Circular}  \\  \hline
    & D-LIE-EKF & MEKF & EULER-EKF  & D-LIE-EKF & MEKF & EULER-EKF  & D-LIE-EKF & MEKF & EULER-EKF\\
    \hline
         Roll &  $0.01861^\circ$ & $0.01911^\circ$ & $0.04610^\circ$ & $0.01672^\circ$ & $0.01666^\circ$ & $0.07520^\circ$ & $0.01730^\circ$ & $0.01740^\circ$ & $0.04206^\circ$\\
         Pitch &  $0.01730^\circ$ & $0.01782^\circ$ & $0.02228^\circ$ & $0.01423^\circ$ & $0.01418^\circ$ & $0.02966^\circ$ & $0.01604^\circ$ & $0.01601^\circ$ & $0.01798^\circ$\\
         Heading & $0.32801^\circ$ & $0.32940^\circ$ & $0.35341^\circ$ & $0.18811^\circ$ & $0.18910^\circ$ & $0.28341^\circ$ & $0.30730^\circ$ & $0.30762^\circ$ & $0.34273^\circ$\\
         Longitude & $0.02246$m & $0.02248$m & $0.02408$m & $0.02240$m & $0.02242$m & $0.02994$m & $0.02241$m & $0.02243$m & $0.02386$m\\
         Latitude & $0.01091$m & $0.01091$m & $0.01134$m & $0.01082$m & $0.01082$m & $0.01131$m & $0.01084$m & $0.01084$m & $0.01120$m\\
         Altitude & $0.00872$m & $0.00876$m & $0.00906$m & $0.00889$m & $0.00893$m & $0.01001$m & $0.00880$m & $0.00883$m & $0.00918$m
    \end{tabular} 
    
    \label{tab:mse.filter}
    }
\end{table}
\begin{table}
\caption{RMSE comparison for 100 Monte Carlo simulations using filter and RTS smoother.}
    \centering
    \resizebox{\columnwidth}{!}{
    \begin{tabular}{c||c|c|c||c|c|c||c|c|c}
    & \multicolumn{3}{c||}{Helicoidal} & \multicolumn{3}{c||}{Rectangular}& \multicolumn{3}{c}{Circular}  \\  \hline
    & D-LIE-EKS & MEKS & EULER-EKS  & D-LIE-EKS & MEKS & EULER-EKS  & D-LIE-EKS & MEKF & EULER-EKS\\
    \hline
         Roll &  $0.00963^\circ$ & $0.01008^\circ$ & $0.02002^\circ$ & $0.00591^\circ$ & $0.00574^\circ$ & $0.02904^\circ$ & $0.00668^\circ$ & $0.00677^\circ$ & $0.01874^\circ$\\
         Pitch &  $0.00929^\circ$ & $0.00986^\circ$ & $0.01216^\circ$ & $0.00539^\circ$ & $0.00521^\circ$ & $0.01882^\circ$ & $0.00635^\circ$ & $0.00640^\circ$ & $0.00786^\circ$\\
         Heading & $0.11881^\circ$ & $0.13661^\circ$ & $0.18123^\circ$ & $0.07431^\circ$ & $0.07796^\circ$ & $0.14142^\circ$ & $0.06981^\circ$ & $0.09176^\circ$ & $0.16601^\circ$\\
         Longitude & $0.00930$m & $0.00931$m & $0.01010$m & $0.00928$m & $0.00929$m & $0.01218$m & $0.00926$m & $0.00927$m & $0.01000$m\\
         Latitude & $0.00428$m & $0.00429$m & $0.00442$m & $0.00423$m & $0.00423$m & $0.00435$m & $0.00428$m & $0.00428$m & $0.00448$m\\
         Altitude & $0.00418$m & $0.00422$m & $0.00446$m & $0.00424$m & $0.00430$m & $0.00467$m & $0.00410$m & $0.00414$m & $0.00445$m
    \end{tabular} }    
    \label{tab:mse.smoother}
\end{table}
One can notice that using the Lie Group-based algorithm the heading error is for the three scenarios lower in terms of RMSE. The D-LIE-EKS shows an overall performance gain of about $40\%$ over the MEKS for both helicoidal and circular flight profiles and about $9\%$ for the rectangular profile. This result indicates the superiority of the Lie Group approach and, more expressively, for curved trajectories.

\subsection{Drone-borne DinSAR performance}
\label{sec:real.data.exp}

\begin{figure}
    \centering
    \includegraphics[width=0.5\textwidth]{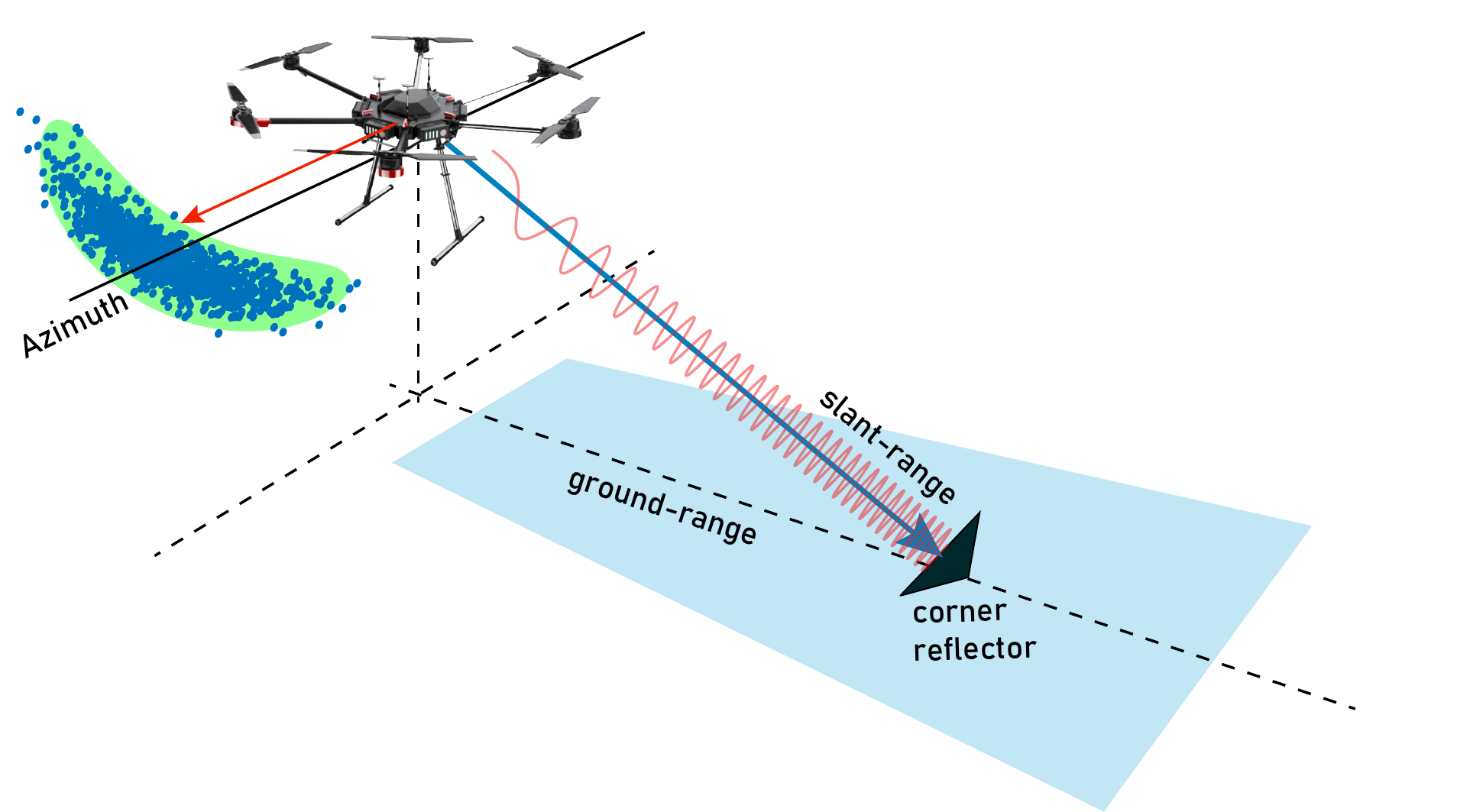}
    \caption{Drone-borne DinSAR geometry.}
    \label{fig:sar}
\end{figure}%
This section reports the performance obtained from a controlled experiment with the drone-borne DinSAR system described in \cite{Moreira2019}.
The digital surface model (DSM) is determined using the cross-track interferometry information provided by the two C-band antennae and applied in the DInSAR calculation.
Three trihedral corner reflectors with square sides and an edge length of 0.6 m were used as a ground reference. Figure \ref{fig:sar} illustrates the drone-borne DinSAR geometry.
\begin{figure*}
    \centering
    \includegraphics[clip,trim=4cm 9.5cm 4cm 9.5cm,width=0.7\textwidth]{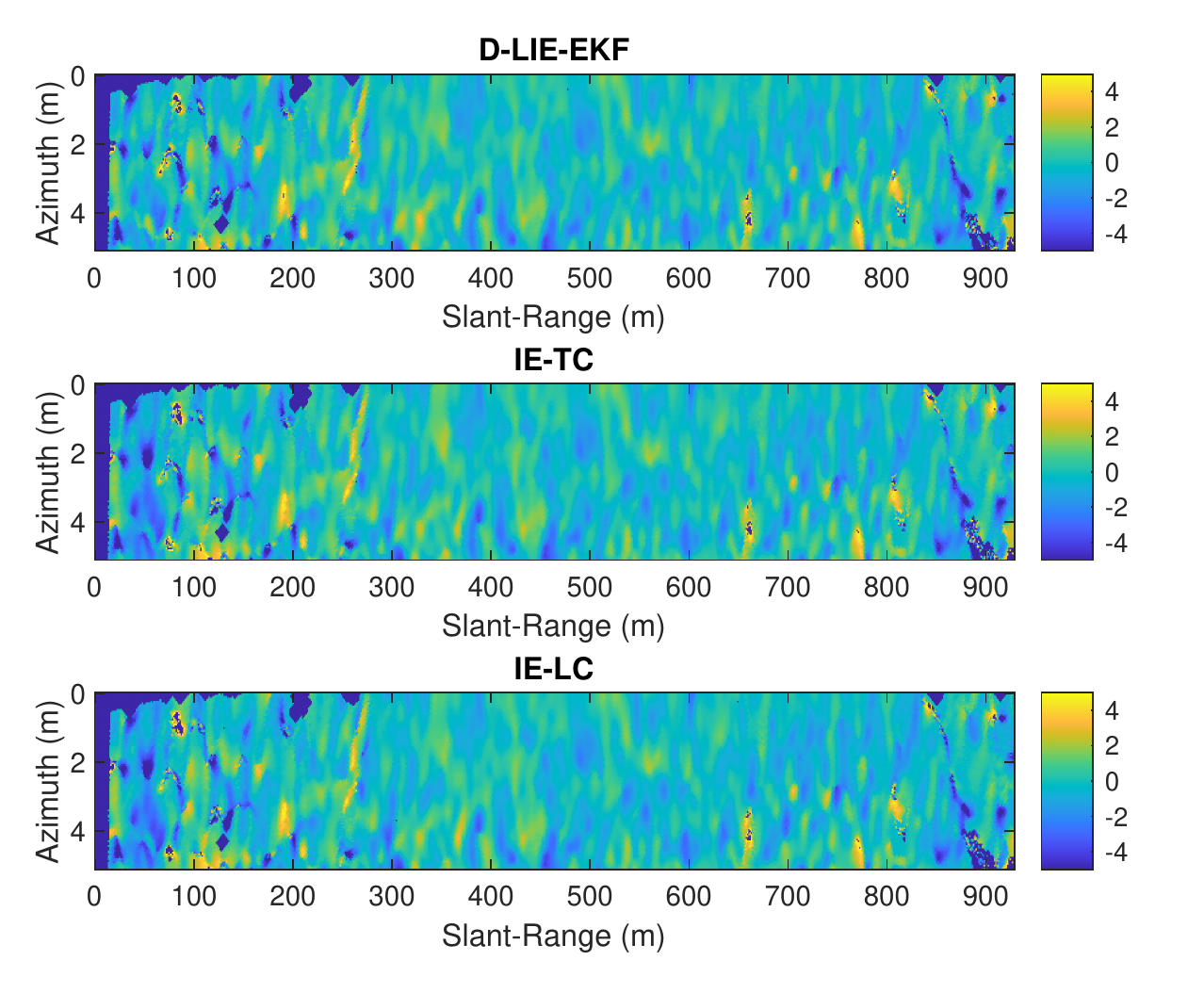}
    \caption{Relative DinSAR Error Image of subsequent flights. For all algorithms, the patterns are almost identical.}
    \label{fig:err.img}
\end{figure*}%
The experiment accurately assesses the DinSAR processing using the proposed GNSS/INS technique and consists of the following steps:
 \begin{itemize}
 \item First, the GNSS ground station is placed close to the starting position of the drone, and the GNSS recording is initiated;
 \item Second, three flights are carried out, each consisting of the following successive steps: turning on the drone and the radar; waiting for 15 min for simultaneous and stationary recording of the ground station and radar GNSS data; taking-off; executing the same west-east flight track; landing; waiting for 15 min for simultaneous and stationary recording of the ground station and radar GNSS data; turning-off the radar and the drone;

 \item Finally, the GNSS ground station and drone are dismounted, and the acquired data is downloaded for post-processing.
 \end{itemize}

For performance comparisons, we test against commercially produced navigation software specialized for IMU-GNSS integration, the NovAtel Inertial Explorer\textregistered ~ (IE). 

After the flights, the data are processed in two steps for the Lie Group-based processing. First, the ground station and rover GNSS receivers are processed using RTKlib\cite{TomojiTakasu2009} to generate centimeter-accuracy position information at 1Hz. Second, the filter-smoother algorithm based on Lie Groups described here generates the final position, velocity, and attitude solution combining the GNSS information with the IMU measurements at 200Hz. We refer to this solution as D-LIE-EKF.

The GNSS and IMU raw data for the IE processing is fed directly to the software. Although our proposed scheme is Loosely Coupled only, we compare it with both the Loosely Coupled as Tightly Coupled solutions of the IE. We refer to these solutions as IE-LC and IE-TC, respectively.

Once the navigation information is available, the raw radar data are processed on the imaging module, recording each resulting single-look complex (SLC) image. After that, the interferometry is performed, yielding the interferogram, topography subtraction, and phase-to-height conversion.

The output consists of two deformation maps plus the three SLC images, all in slant range geometry. Each interferogram is calculated with $0.047$m resolution in azimuth and $1.228$m resolution in the slant range. 

From the close inspection of the reflectors positioning, we can not detect any noticeable advantages from one or the other procedure. After these evaluations, we attempted to spot subtle differences by subtracting images of subsequent flights and looking for terrain inconsistencies. 

Figure \ref{fig:err.img} shows the relative error of subsequent flights for the three different navigation processing. Notice that the resulting pattern is almost identical for all algorithms. 

\begin{figure}
    \centering
    \includegraphics[clip,trim=3cm 11.5cm 4cm 12cm,width=0.45\textwidth]{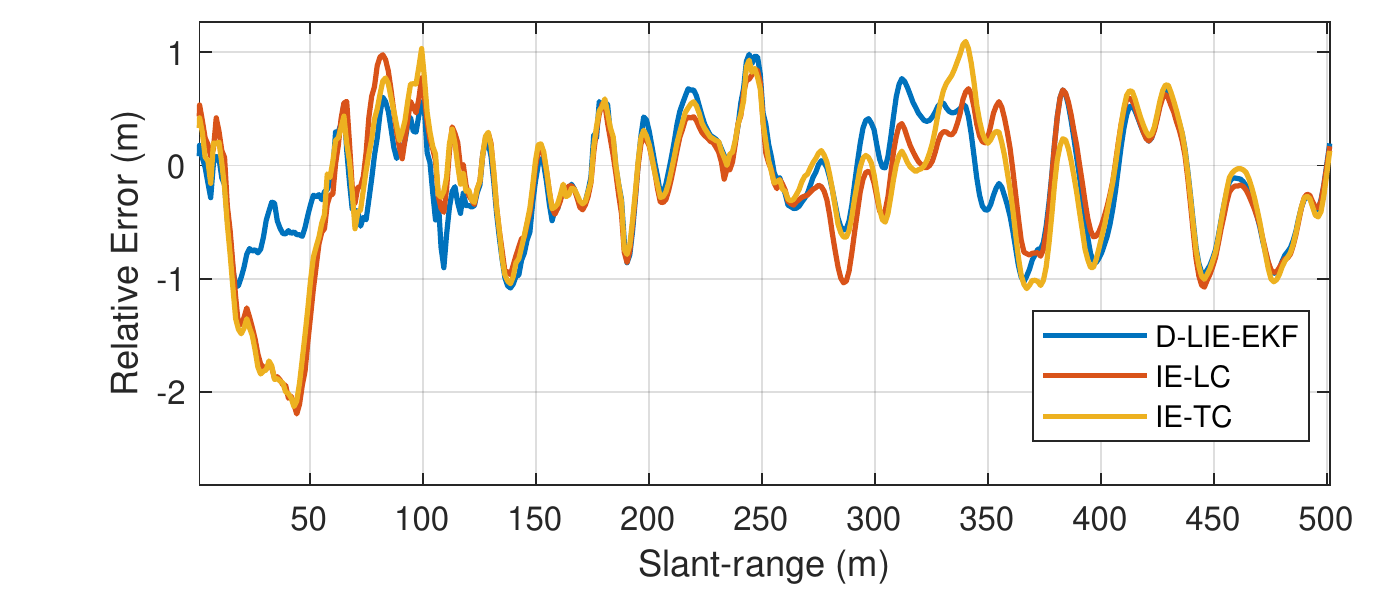}
    \caption{Relative error of DinSAR for one azimuth line.}
    \label{fig:err.line}
\end{figure}

Figure \ref{fig:err.line} depicts the relative error for only one azimuth line (horizontal cross-section of Figure \ref{fig:err.img}). Between slant-range $0-100$m, one can observe an increase in relative error for IE-LC and IE-TC, while for the D-LIE-EKF, the error magnitude remains steady.

Table \ref{tbl:dinsar.rmse} summarizes the overall error of the three algorithms. We observe that the proposed D-LIE-EKF scheme outperforms both IE-LC and IE-TC in terms of RMSE.
\begin{table}
\begin{center}
\caption{RMSE comparison of the DinSAR relative error.}
\label{tbl:dinsar.rmse}
    \begin{tabular}{c||c|c|c}
   & D-LIE-EKF & IE-LC & IE-TC \\  \hline
    RMSE &  $1.136\mbox{m}$ & $1.179\mbox{m}$ & $1.174\mbox{m}$ 
    \end{tabular} 
    \end{center}
\end{table}%
\section{Conclusion}
\label{sec:conclusion}
A novel scheme for loosely coupled GNSS/INS integration was developed based on Lie Group Theory, tailored for post-processing applications and seeking the highest accuracy. It combines the online Kalman Filter with the offline RTS smoother to provide estimates of PVA to this goal. The double direct isometry group $\SE_2(3)$ compound with the translation group $T(6)$ was adopted to model the kinematic and cope with biased navigation system measurements.

The paper presents some numeric experiments based on synthetic data generated using inverse strap-down mechanization. They show that the Lie Group approach consistently outperforms classical approaches based on quaternion and Euler parametrization of the attitude matrix. The advantage of the Lie Group is further stressed using helicoidal and circular trajectories in which the proposed scheme attains better RMS performance.

Furthermore, in a field comparison, the Lie Group also exhibits superior performance when applied to a DinSAR drone-borne application and shows better accuracy than commercially available navigation software Inertial Explorer\textregistered. Surprisingly, the simple Loosely Coupled configuration of the proposed technique yielded superior performance than the IE's more complex Tightly Coupled configuration.

In conclusion, this work shows that the Lie Group theory's filtering and smoothing are applicable for post-processing loosely coupled GNSS/INS integration. It can yield performance improvement for aerial navigation, making it an appealing framework for applications that require high-quality navigation information, such as DinSAR. Finally, this work also points out that the air-borne radar scheme employed here is excellent for benchmark GNSS/INS integration.

\section{Acknowledgements}
The authors are thankful to the Radaz Indústria e
Comércio de Produtos Eletrônicos S.A. for supporting this work.

\appendix

\subsection{D-LIE-EKF}
\label{app:d.lie.ekf}
\begin{proof}{\cref{lemma:d.ekf}.}
Let $X_k=\hat{X}_{k|k}\exp_G^\wedge(\epsilon_{k|k})$ with $\epsilon_{k|k}\sim\mathcal{N}(0,P_{k|k})$ and $\hat{X}_{k+1|k}=\hat{X}_{k|k}\exp_G^\wedge(\hat\Omega_k)$. Employing a first-order approximation for $\Omega_k$ around $\hat{X}_{k|k}$, yields
\begin{equation}
    \Omega_k=\Omega(\hat{X}_{k|k}\exp_G^\wedge(\epsilon_{k|k}))\approxeq \Omega(\hat{X}_{k|k})+\mathscr{C}_k\epsilon_{k|k}
\end{equation}
where $\mathscr{C}_k:=\frac{\partial}{\partial \epsilon}[\Omega(\hat{X}_{k|k}\exp_G^\wedge(\epsilon))]\Big|_{\epsilon=0}$.
From \cref{eq:sys.lie}, one has,
\begin{multline}
    \epsilon_{k+1|k}=\hat{X}_{k+1|k}^{-1}X_{k+1}=\exp_G^\wedge(-\hat\Omega_k)\hat{X}_{k|k}^{-1}X_k\exp_G^\wedge(\Omega_k+w_k)\\=\exp_G^\wedge(-\hat\Omega_k)\exp_{G}^\wedge(\epsilon_{k|k})\exp_G^\wedge(\hat{\Omega}_k+\mathscr{C}_k\epsilon_{k|k}+w_k)
\end{multline}

Assuming that $\mathscr{C}_k\epsilon_{k|k}+w_k$ is small, then using the relationship from
\cite{Barfoot2019} and the fact that $g\exp_G^\wedge(x) = \exp_G^\wedge\left( \Ad_G(g)x\right)g$ one gets
\begin{equation}
     \epsilon_{k+1|k}=\exp_{G}^\wedge(\mathscr{F}\epsilon_{k|k}+J_r(\hat{\Omega}_k)w_k)
\end{equation}
with $\mathscr{F}=\Ad_G(\exp_G^\wedge(-\hat{\Omega}_k))+J_r(\hat{\Omega}_k)\mathscr{C}_k$.
Therefore, we conclude that $X_{k+1}\sim\mathscr{N}_G(\hat{X}_{k+1|k},P_{k+1|k})$ where
\begin{subequations}
\begin{align}    P_{k+1|k}&=\mathscr{F}P_{k|k}\mathscr{F}^\trp+J_r(\hat{\Omega}_k)Q_kJ_r(\hat{\Omega}_k)^\trp.
\end{align}
\end{subequations}
Now, for the measurement-update step, consider
$X_{k+1}=\hat{X}_{k+1|k}\exp_G^\wedge(\epsilon_{k+1|k})$
 where $\epsilon_{k+1|k}\sim\mathcal{N}(0,P_{k+1|k})$.
 Note that the measurement distribution is
$
    p(y_{k+1}|X_{k+1})=\mathcal{N}(y_{k+1};h(X_{k+1}),R_{k+1})
$ and the prior state distribution is $p(X_{k+1}|y_{1:k})=\mathscr{N}_G(\hat{X}_{k+1|k},P_{k+1|k})$. Thus, from the Bayes' rule, one has
\begin{multline}
    p(X_{k+1}|y_{1:k+1})=\frac{p(y_{k+1}|X_{k+1})p(X_{k+1}|y_{1:k})}{p(y_{k+1}|y_{1:k})}\\\propto \exp\left(-\frac{1}{2}\|y_{k+1}-h(X_{k+1})\|_{R_{k+1}^{-1}}^2 -\frac{1}{2}\|\epsilon_{k+1|k}\|_{P_{k+1|k}^{-1}}^2\right).
\end{multline}
Let the posterior distribution be parametrized in the form
\begin{equation}
    p(X_{k+1}|y_{1:k+1})=p(\hat{X}_{k+1|k}\exp_G^\wedge(v)|y_{1:k+1}).
\end{equation}
Therefore, the Maximum A Posteriori (MAP) estimate is $\hat{X}_{k+1|k+1}=\hat{X}_{k+1|k}\exp_G^\wedge(v^*)$ such that $v^*=\arg\min_v \ell(v)$ where $\ell(v)$ is the negative log-likelihood given by,
\begin{equation*}
    \ell(v):=\frac{1}{2}\|y_{k+1}- h(\hat{X}_{k+1|k}\exp_G^\wedge(v))\|_{R_{k+1}^{-1}}^2+
    \frac{1}{2}\|v\|_{P_{k+1|k}^{-1}}^2
\end{equation*}
If a linear approximation is adopted, then one can obtain a set of filtering equations similar to the Extended Kalman Filter (EKF). For this purpose, consider a first-order approximation of \cref{eq:measurement.lie} in the form
$
     h(\hat{X}_{k+1|k}\exp_G^\wedge(\epsilon))\approxeq h(\hat{X}_{k+1|k})+\mathscr{H}\epsilon
$
 with $\mathscr{H}=\frac{\partial}{\partial \epsilon}[h(\hat{X}_{k+1|k}\exp_G^\wedge(\epsilon))]\Big|_{\epsilon=0}$. Thus,
 \begin{equation}
 \label{eq:ell.linear}
    \ell(v)\approxeq\frac{1}{2}\|z_{k+1}- \mathscr{H}v\|_{R_{k+1}^{-1}}^2+
    \frac{1}{2}\|v\|_{P_{k+1|k}^{-1}}^2
\end{equation}
where $z_{k+1}:=y_{k+1}-h(\hat{X}_{k+1|k})$ is the residual. The minimum of \cref{eq:ell.linear} is straightforward given by $v^*=Kz_{k+1}$ where, $K=(P_{k+1|k}^{-1}+\mathscr{H}^\trp R_{k+1}^{-1}\mathscr{H})^{-1}\mathscr{H}^\trp R_{k+1}^{-1}$ which also can be written as $K=P_{k+1|k}\mathscr{H}^\trp(R_{k+1}+\mathscr{H}P_{k+1|k}\mathscr{H}^\trp)^{-1}$. Thus,
\begin{align}
\label{eq:posterior.state}
    \hat{X}_{k+1|k+1}&=\hat{X}_{k+1|k}\exp_G^\wedge(Kz_{k+1}).
\end{align}
Let the state error be $\epsilon_{k+1|k+1}:=\log_G^\vee(\hat{X}_{k+1|k+1}^{-1}X_{k+1})$. From \cref{eq:posterior.state}, the posterior error follows,
\begin{equation}
\begin{aligned}
    \epsilon_{k+1|k+1}&=\log_G^\vee(\hat{X}_{k+1|k+1}^{-1}X_{k+1})
    =\log_G^\vee(\exp_G^\wedge(-Kz_{k+1})\hat{X}_{k+1|k}^{-1}X_{k+1})\\
   &=\log_G^\vee(\exp_G^\wedge(-Kz_{k+1})\exp_G^\wedge(\epsilon_{k+1|k}))
   =-K(y_{k+1}-h(\hat{X}_{k+1|k}))+\epsilon_{k+1|k}\\
    &\approx-K(\mathscr{H}\epsilon_{k+1|k}+\nu_{k+1})+\epsilon_{k+1|k}
    =(I-K\mathscr{H})\epsilon_{k+1|k}-K\nu_{k+1}
\end{aligned}
\end{equation}
where a linear approximation  $z_{k+1}\approx \mathscr{H}\epsilon_{k+1|k}+\nu_{k+1}$ is employed. Define $P_{k+1|k+1}:=\mathbb{E}[\epsilon_{k+1|k+1}\epsilon_{k+1|k+1}^\trp]$. Assuming that $\mathbb{E}[\epsilon_{k+1|k}\nu_{k+1}^\trp]=0$, one gets,
\begin{equation}
\label{eq:posterior.cov}
    P_{k+1|k+1}=(I-K\mathscr{H})P_{k+1|k}(\bullet)^\trp + KR_{k+1}K^\trp.
\end{equation}
\end{proof}

\subsection{RTS on Lie Groups}
\label{app:rts.lie}
\begin{proof}{\cref{lemma:d.eks}.}
Following the lines of \cite{Sarkka2013}, from a Bayesian perspective, the RTS smoother on Lie Groups can be derived as the Maximum a Posteriori (MAP) estimate of the following joint pdf 
\begin{multline}
\label{eq:rts.pdf}
 p(X_k,X_{k+1}|y_{1:T})=p(X_{k}|X_{k+1},y_{1:k})p(X_{k+1}|y_{1:T})\\
 =\frac{p(X_{k+1}|X_{k})p(X_k|y_{1:k})}{p(X_{k+1}|y_{1:k})}p(X_{k+1}|y_{1:T}) 
 =\frac{p(X_{k+1},X_{k}|y_{1:k})}{p(X_{k+1}|y_{1:k})}p(X_{k+1}|y_{1:T}).
\end{multline}
Assume that 
\begin{subequations}
\begin{align}
    p(X_{k+1},X_k|y_{1:k})&=\mathscr{N}_{G\times G}\left(\hat{X}_{k|k},\hat{X}_{k+1|k},\mathcal{P}
    \right)\\ 
    p(X_{k}|y_{1:k})&=\mathscr{N}_G(\hat{X}_{k|k},P_{k|k})\\ 
    p(X_{k+1}|y_{1:T})&=\mathscr{N}_G(\hat{X}_{k+1}^s,P_{k+1}^s)\\ 
    p(X_{k+1}|y_{1:k})&=\mathscr{N}_G(\hat{X}_{k+1|k},P_{k+1|k})
    \end{align}
    \end{subequations}
where $\mathcal{P}=\begin{bmatrix} 
    P_{k|k} & P_{k|k}\mathscr{F}_k^\trp \\ 
    \mathscr{F}_kP_{k|k} & P_{k+1|k}
    \end{bmatrix}$ and $\{\hat{X}_{k|k},P_{k|k}\}_{k=1:N}$ comes from the D-LIE-EKF.  Accordingly, \cref{eq:rts.pdf} becomes
\begin{multline}
 p(X_k,X_{k+1}|y_{1:T})
 \propto \exp\Bigl(-\frac{1}{2}\left\|\begin{bmatrix} 
\log_G^\vee[\hat{X}_{k|k}^{-1}X_k]\\ 
 \log_G^\vee[\hat{X}_{k+1|k}^{-1}X_{k+1}]
 \end{bmatrix}
 \right\|^2_{\mathcal{P}^{-1}}
 \\+\frac{1}{2}\|\log_G^\vee[\hat{X}_{k+1|k}^{-1}X_{k+1}]\|^2_{P_{k+1|k}^{-1}} 
 -\frac{1}{2}\|\log_G^\vee[(\hat{X}_{k+1}^s)^{-1}X_{k+1}]\|^2_{(P_{k+1}^s)^{-1}}
 \Bigr).
 \label{eq:rts.pdf.cgd}
\end{multline}
Parametrize $X_k=\hat{X}_{k|k}\exp_G^\wedge(\delta_k)$ and $X_{k+1}=\hat{X}_{k+1}^s\exp_G^\wedge(\delta_{k+1})$.
Thereafter, taking the negative logarithm of \cref{eq:rts.pdf.cgd} yields
\begin{multline}
 \ell(\delta_k,\delta_{k+1})
 = \frac{1}{2}\left\|\begin{bmatrix} 
 \delta_k\\ 
 \log_G^\vee[\exp_G^\wedge(z_s)\exp_G^\wedge(\delta_{k+1})]
 \end{bmatrix}
 \right\|^2_{\mathcal{P}^{-1}}
 \\-\frac{1}{2}\| \log_G^\vee[\exp_G^\wedge(z_s)\exp_G^\wedge(\delta_{k+1})]\|^2_{P_{k+1|k}^{-1}} 
 +\frac{1}{2}\|\delta_{k+1}\|^2_{(P_{k+1}^s)^{-1}}
\end{multline}
where $z_s:=\log_G^\vee[\hat{X}_{k+1|k}^{-1}\hat{X}_{k+1}^s]$. Assuming that both $z_s$ and $\delta_{k+1}$ are small, one has
\begin{multline}
 \ell(\delta_k,\delta_{k+1})
 =\\ \frac{1}{2}\left\|\begin{bmatrix} 
 \delta_k\\ 
 z_s+\delta_{k+1}
 \end{bmatrix}
 \right\|^2_{\mathcal{P}^{-1}}
 -\frac{1}{2}\| z_s+\delta_{k+1}\|^2_{P_{k+1|k}^{-1}} 
 +\frac{1}{2}\|\delta_{k+1}\|^2_{(P_{k+1}^s)^{-1}}
 =\frac{1}{2}\left\|\mathcal{A}\delta+b
 \right\|^2_{\mathcal{W}}
  \label{eq:rts.cost}
\end{multline}
where 
\begin{equation}
    \delta:=\begin{bmatrix} 
 \delta_k\\ 
 \delta_{k+1}
 \end{bmatrix},\quad\mathcal{A}:=\begin{bmatrix}
    I & 0\\ 
    0 & I\\ 
    0 & I\\ 
    0 & I
    \end{bmatrix},\quad b:=\begin{bmatrix}
    0\\ 
    z_s\\ 
    z_s\\ 
    0
    \end{bmatrix} 
\end{equation}
and $\mathcal{W}=\blkdiag(\mathcal{P}^{-1},-P_{k+1|k}^{-1},(P_{k+1}^s)^{-1})$. The optimal solution of \cref{eq:rts.cost} is straightforward given by 
\begin{equation}
\begin{aligned}
    \delta^*&=(\mathcal{A}^\trp \mathcal{W}\mathcal{A})^{-1}\mathcal{A}^\trp \mathcal{W}b
    =\begin{bmatrix}
    \delta_k^*\\
    \delta_{k+1}^*
    \end{bmatrix}=\begin{bmatrix}
    G_kz_s\\
    0
    \end{bmatrix}.
    \end{aligned}
\end{equation}
Finally, the smoothed state estimate is $\hat{X}_k^s=\hat{X}_{k|k}\exp_G^\wedge(G_kz_s)$ and the final covariance is obtained from the first block diagonal of $(\mathcal{A}^\trp \mathcal{W}\mathcal{A})^{-1}$ (see \cite{Bourmaud2015}).
\end{proof}
\bibliography{references.bib}

\end{document}